\documentclass[10pt,english,journal,compsoc]{IEEEtran}
\usepackage[T1]{fontenc}
\usepackage[latin9]{inputenc}
\usepackage{color}
\usepackage{array}
\usepackage{bm}
\usepackage{multirow}
\usepackage{amsmath}
\usepackage{amssymb}
\usepackage{graphicx}

\makeatletter

\providecommand{\tabularnewline}{\\}

\usepackage{graphicx}
\usepackage{amsmath,amssymb} % define this before the line numbering.

\usepackage{color}
\usepackage{array}
\usepackage{multirow}
\usepackage{amsmath}
\usepackage{graphicx}
\usepackage{color}
\usepackage{url}

\makeatother

\usepackage{babel}
\begin{document}

\title{Transductive Multi-view Zero-Shot Learning} 
\author{Yanwei~Fu,~\IEEEmembership{} 
Timothy~M.~Hospedales,~ Tao~Xiang~ and Shaogang~Gong 
\IEEEcompsocitemizethanks{ \IEEEcompsocthanksitem Yanwei Fu is with Disney Research, Pittsburgh, PA, 15213, USA.
\IEEEcompsocthanksitem Timothy~M.~Hospedales, Tao~Xiang,   and Shaogang~Gong are with the School of Electronic Engineering and Computer Science, Queen Mary University of London, E1 4NS, UK.\protect\\ 
Email: \{y.fu,t.hospedales,t.xiang,s.gong\}@qmul.ac.uk 
}\thanks{} }

%\markboth{Submission to IEEE Transactions on Pattern Analysis and Machine Intelligence}{} 
\IEEEcompsoctitleabstractindextext{ 
\begin{abstract}
Most existing zero-shot learning approaches exploit transfer learning via an intermediate semantic representation shared between an annotated auxiliary dataset and a target dataset with different classes and no annotation. A projection from a low-level feature space to the semantic representation space is learned from the auxiliary dataset and applied without adaptation to the target dataset. In this paper we identify two inherent limitations with these approaches. First, due to having disjoint and potentially unrelated classes, the projection functions learned from the auxiliary dataset/domain are biased when applied directly to the target dataset/domain.  We call this problem the \textit{projection domain shift} problem and propose a novel framework, {\em transductive multi-view embedding}, to solve it.  The second limitation is the \textit{prototype sparsity} problem which refers to the fact that for each target class, only a single prototype is available for zero-shot learning given a semantic representation.  To overcome this problem, a novel  heterogeneous multi-view hypergraph label propagation method is formulated for zero-shot learning in the transductive embedding space. It effectively exploits the complementary information offered by different semantic representations and takes advantage of the manifold structures of multiple representation spaces in a coherent manner. We demonstrate through extensive experiments that the proposed approach  (1) rectifies the projection shift between the auxiliary and target domains, (2) exploits the complementarity of multiple semantic representations, (3) significantly outperforms existing methods for both zero-shot and N-shot recognition on three image and video benchmark datasets, and (4) enables novel cross-view annotation tasks.

\end{abstract} 
 } \maketitle

\section{Introduction}

Humans can distinguish 30,000 basic object classes \cite{object_cat_1987}
and many more subordinate ones (e.g.~breeds of dogs). They can also
create new categories dynamically from few examples or solely based
on high-level description. In contrast, most existing computer vision
techniques require hundreds of labelled samples for each object class
in order to learn a recognition model. Inspired by humans' ability
to recognise without seeing samples, and motivated by the prohibitive
cost of training sample collection and annotation, the research area
of \emph{learning to learn} or \emph{lifelong learning} \cite{PACbound2014ICML,chen_iccv13}
has received increasing interests. These studies aim to intelligently
apply previously learned knowledge to help future recognition tasks.
In particular, a major and topical challenge in this area is to build
recognition models capable of recognising novel visual categories
without labelled training samples, i.e.~zero-shot learning (ZSL).

\begin{figure}[t]
\centering{}\includegraphics[scale=0.26]{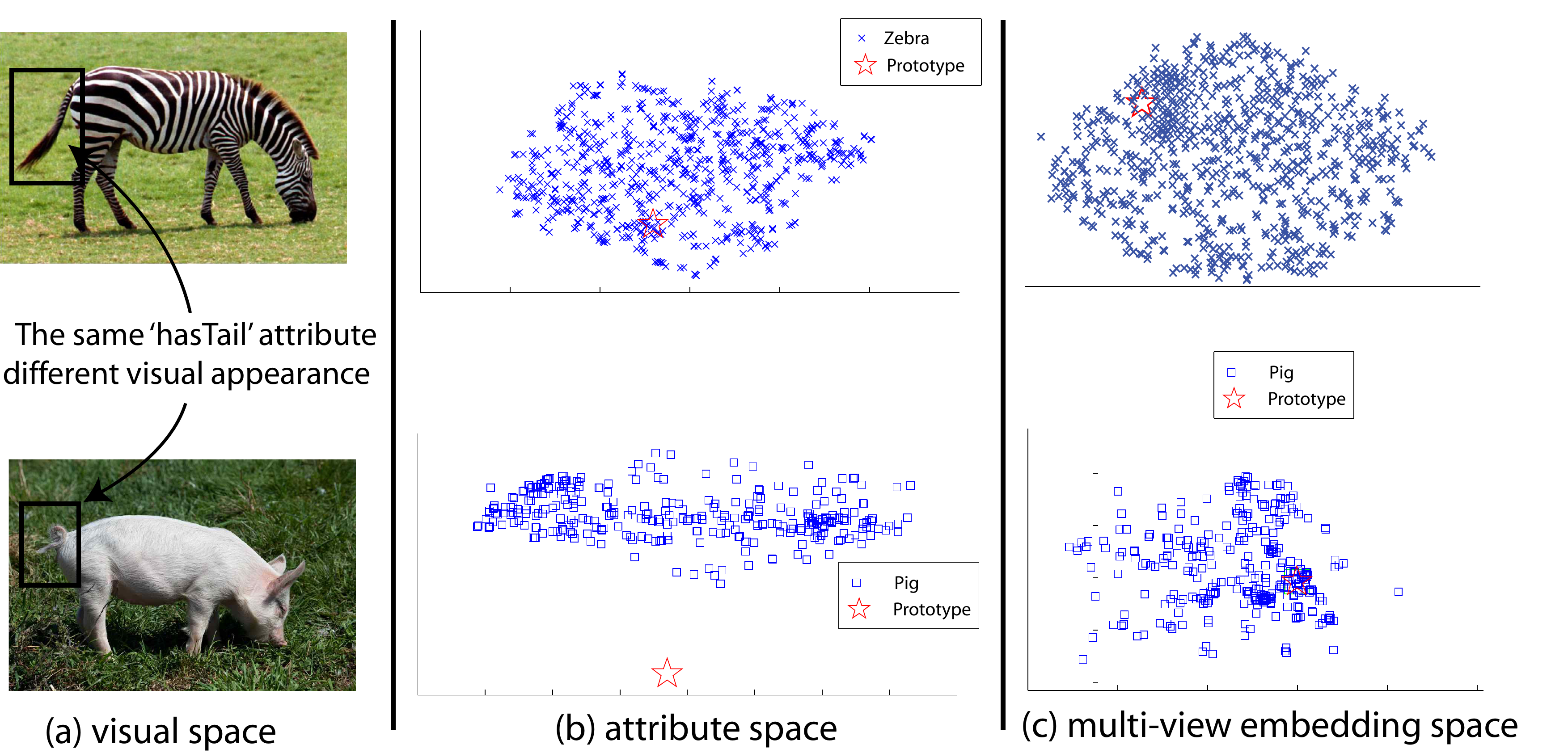}\caption{\label{fig:domain-shift:Low-level-feature-distribution}An illustration
of the projection domain shift problem. Zero-shot prototypes are shown
as red stars and predicted semantic attribute projections
(defined in Sec.~3.2) shown in blue.}
\end{figure}

The key idea underpinning  ZSL approaches is to exploit knowledge
transfer via an intermediate-level semantic representation. Common
semantic representations include binary vectors of visual attributes
\cite{lampert2009zeroshot_dat,liu2011action_attrib,yanweiPAMIlatentattrib}
(e.g. 'hasTail' in Fig.~\ref{fig:domain-shift:Low-level-feature-distribution})
and continuous word vectors \cite{wordvectorICLR,DeviseNIPS13,RichardNIPS13}
encoding linguistic context. In ZSL, two datasets with disjoint classes
are considered: a labelled auxiliary set where a semantic representation
is given for each data point, and a target dataset to be classified
without any labelled samples. The semantic representation is assumed
to be shared between the auxiliary/source and target/test
dataset. It  can thus be re-used for knowledge transfer between the source and
target sets: a projection function mapping
low-level features to the semantic representation is learned from
the auxiliary data by classifier or regressor.
This projection is then applied to map each unlabelled
target class instance into the same semantic space. 
In this space, a `prototype' of each target class is specified, and each projected target instance is classified
by measuring similarity to the class prototypes. 
Depending on the semantic space, the class prototype could be a binary attribute
vector listing class properties (e.g., 'hasTail') \cite{lampert2009zeroshot_dat}
or a word vector describing the linguistic context of the textual
class name \cite{DeviseNIPS13}.

Two inherent problems exist in this conventional zero-shot learning
approach. The first problem is the \textbf{projection domain shift
problem}. Since the two datasets have different and potentially unrelated
classes, the underlying data distributions of the classes differ,
so do the `ideal' projection functions between the low-level feature
space and the semantic spaces. Therefore, using the projection functions
learned from the auxiliary dataset/domain without any adaptation to
the target dataset/domain causes an unknown shift/bias. We call it
the \textit{projection domain shift} problem. This is illustrated
in Fig.~\ref{fig:domain-shift:Low-level-feature-distribution}, which
shows two object classes from the Animals with Attributes (AwA) dataset
\cite{lampert13AwAPAMI}: Zebra is one of the 40 auxiliary classes
while Pig is one of 10 target classes. Both of them share the same
`hasTail' semantic attribute, but the visual appearance of their tails
differs greatly (Fig.~\ref{fig:domain-shift:Low-level-feature-distribution}(a)).
Similarly, many other attributes of Pig are visually different from
the corresponding attributes in the auxiliary classes. Figure \ref{fig:domain-shift:Low-level-feature-distribution}(b)
illustrates the projection domain shift problem by plotting (in 2D
using t-SNE \cite{tsne}) an 85D attribute space representation of
the image feature projections and class prototypes (85D binary attribute
vectors). A large discrepancy can be seen between the Pig prototype
in the semantic attribute space and the projections of its class member
instances, but not for the auxiliary Zebra class. This discrepancy
is caused when the projections learned from the 40 auxiliary classes
are applied directly to project the Pig instances -- what `hasTail'
(as well as the other 84 attributes) visually means is different now.
Such a discrepancy will inherently degrade the effectiveness of zero-shot
recognition of the Pig class because the target class instances are
classified according to their similarities/distances to those prototypes.
To our knowledge, this problem has neither been identified nor addressed
in the zero-shot learning literature.

The second problem is the \textbf{prototype sparsity problem}: for
each target class, we only have a single prototype which is insufficient
to fully represent what that class looks like. As shown in Figs.~\ref{fig:t-SNE-visualisation-of}(b)
and (c), there often exist large intra-class variations and inter-class
similarities. Consequently, even if the single prototype is centred
among its class instances in the semantic representation space, existing
zero-shot classifiers will still struggle to assign correct class
labels -- one prototype per class is not enough to represent the intra-class
variability or help disambiguate class overlap \cite{Eleanor1977}. % This problem has not been explicitly identified\textcolor{red}{{} in computer vision community}% \footnote{\textcolor{red}{It has been considered in prototype theory, studied by psychophysicists in the 1970s \cite{Eleanor1977}.}% } although a partial solution exists \cite{transferlearningNIPS}. 

In addition to these two  problems, conventional approaches to
zero-shot learning are also limited in \textbf{exploiting multiple
intermediate semantic representations}. Each representation (or semantic
`view') may contain complementary information -- useful for
distinguishing different classes in different ways. While both visual attributes \cite{lampert2009zeroshot_dat,farhadi2009attrib_describe,liu2011action_attrib,yanweiPAMIlatentattrib}
and linguistic semantic representations such as word vectors \cite{wordvectorICLR,DeviseNIPS13,RichardNIPS13}
have been independently exploited successfully, it remains unattempted
and non-trivial to synergistically exploit  multiple semantic
views. This is because they are often of very different dimensions
and types and each suffers from different domain shift effects discussed
above. %This exploitation has to be transductive for zero-shot learning as only unlabelled data are available for the target classes and the labelled auxiliary data cannot be used directly due to the projection domain shift problem.

%our solution the first problem is to fusing multiple semantic views in a transductive embedding framework. Why transductive, why multiple view. why embedding low-level feature. the more view you fuse, the better aligned, and importantly the data of different class become more seperable. We use CCA.

In this paper, we propose to solve the projection domain shift problem
using transductive multi-view embedding. The
transductive setting means using the unlabelled test data to improve
generalisation accuracy. In our framework, each unlabelled
target class instance is represented by multiple views: its low-level
feature view and its (biased) projections in multiple semantic spaces
(visual attribute space and word space in this work). To rectify the
projection domain shift between auxiliary and target datasets, we
introduce a multi-view semantic space alignment process to correlate
different semantic views and the low-level feature view by projecting
them onto a common latent embedding space learned using multi-view Canonical
Correlation Analysis (CCA) \cite{multiviewCCAIJCV}. The intuition is that when the biased target data projections (semantic representations) are correlated/aligned with their (unbiased) low-level feature representations, the bias/projection domain shift is alleviated.  The effects of this process
on projection domain shift are illustrated by Fig.~\ref{fig:domain-shift:Low-level-feature-distribution}(c),
where after alignment, the target Pig class prototype is much closer
to its member points in this embedding space. Furthermore, after exploiting
the complementarity of different low-level feature and semantic views
synergistically in the common embedding space, different target classes
become more compact and more separable (see Fig.~\ref{fig:t-SNE-visualisation-of}(d)
for an example), making the subsequent zero-shot recognition a much
easier task.

Even with the proposed transductive multi-view embedding framework,
the prototype sparsity problem remains -- instead of one prototype
per class, a handful are now available depending on how many views
are embedded, which are still sparse. Our solution
is to pose this as a semi-supervised learning \cite{zhu2007sslsurvey}
problem:  prototypes in each view are treated as labelled `instances',
and we exploit the manifold structure of the unlabelled data distribution
in each view in the embedding space via label propagation on a graph.
To this end, we introduce a novel transductive multi-view hypergraph
label propagation (TMV-HLP) algorithm for recognition. The
core in our TMV-HLP algorithm is a new {\emph{distributed
representation} of graph structure termed heterogeneous
hypergraph which allows us to exploit the complementarity of different
semantic and low-level feature views, as well as the manifold structure
of the target data to compensate for the impoverished supervision
available from the sparse prototypes. Zero-shot learning is then
performed by semi-supervised label propagation from the prototypes
to the target data points within and across the graphs. The whole
framework is illustrated in Fig.~\ref{fig:The-pipeline-of}.

By combining our transductive embedding framework and the TMV-HLP
zero-shot recognition algorithm, our approach generalises seamlessly
when none (zero-shot), or few (N-shot) samples of the target classes
are available. Uniquely it can also synergistically exploit zero +
N-shot (i.e., both prototypes and labelled samples) learning. Furthermore,
the proposed method enables a number of novel cross-view annotation
tasks including \textit{zero-shot class description} and \textit{zero
prototype learning}.

\noindent \textbf{Our contributions}\quad{}Our contributions are
as follows: (1) To our knowledge, this is the first attempt to investigate
and provide a solution to the projection domain shift problem in zero-shot
learning. (2) We propose a transductive multi-view embedding space
that not only rectifies the projection shift, but also exploits the
complementarity of multiple semantic representations of visual data.
(3) A novel transductive multi-view heterogeneous hypergraph label
propagation algorithm is developed to improve both zero-shot and N-shot
learning tasks in the embedding space and overcome the prototype sparsity
problem. (4) The learned embedding space enables a number of novel
cross-view annotation tasks. Extensive experiments are carried
out and the results show that our approach significantly outperforms
existing methods for both zero-shot and N-shot recognition on three
image and video benchmark datasets.

\section{Related Work}

\noindent \textbf{Semantic spaces for zero-shot learning}\quad{}To
address zero-shot learning, attribute-based semantic representations have
been explored for images \cite{lampert2009zeroshot_dat,farhadi2009attrib_describe}
and to a lesser extent videos \cite{liu2011action_attrib,yanweiPAMIlatentattrib}.
Most existing studies \cite{lampert2009zeroshot_dat,hwang2011obj_attrib,palatucci2009zero_shot,parikh2011relativeattrib,multiattrbspace,Yucatergorylevel,labelembeddingcvpr13}
assume that an exhaustive ontology of attributes has been manually
specified at either the class or instance level. However, annotating
attributes scales poorly as ontologies tend to be domain specific.
This is despite efforts exploring augmented data-driven/latent attributes
at the expense of name-ability \cite{farhadi2009attrib_describe,liu2011action_attrib,yanweiPAMIlatentattrib}.
To address this, semantic representations using existing ontologies
and incidental data have been proposed \cite{marcuswhathelps,RohrbachCVPR12}.
Recently, \emph{word vector} approaches based on distributed language
representations have gained popularity. In this case a word space
is extracted from linguistic knowledge bases e.g.,~Wikipedia by natural
language processing models such as \cite{NgramNLP,wordvectorICLR}.
The language model is then used to project each  class'
textual name into this space. These projections can be used as  prototypes for zero-shot learning
\cite{DeviseNIPS13,RichardNIPS13}. Importantly, regardless of the
semantic spaces used, existing methods focus on either designing better
semantic spaces or how to best learn the projections. The former is
orthogonal to our work -- any semantic spaces can be used in our framework
and better ones would benefit our model. For the latter, no existing
work has identified or addressed the projection domain shift problem.

\noindent \textbf{Transductive zero-shot learning} 
 was considered by Fu et al.~\cite{fu2012attribsocial,yanweiPAMIlatentattrib}
who introduced a generative model to for user-defined and latent
attributes. A simple transductive zero-shot learning algorithm is
proposed: averaging the prototype's k-nearest neighbours to exploit
the test data attribute distribution. Rohrbach
et al.~\cite{transferlearningNIPS}
proposed a more elaborate transductive strategy, using graph-based
label propagation to exploit the manifold structure of the test data.
These studies effectively transform the ZSL task into a transductive
semi-supervised learning task \cite{zhu2007sslsurvey} with prototypes
providing the few labelled instances. Nevertheless,
 these studies and this paper (as with most previous work \cite{lampert13AwAPAMI,lampert2009zeroshot_dat,RohrbachCVPR12})
only consider recognition among the novel classes: unifying zero-shot
with supervised learning remains an open challenge \cite{RichardNIPS13}.

\noindent \textbf{Domain adaptation}\quad{}Domain adaptation methods
attempt to address the domain shift problems that occur when the assumption
that the source and target instances are drawn from the same distribution
is violated. Methods have been derived for both classification \cite{fernando2013unsupDAsubspace,duan2009transfer}
and regression \cite{storkey2007covariateShift}, and both with \cite{duan2009transfer}
and without \cite{fernando2013unsupDAsubspace} requiring label information
in the target task. Our zero-shot learning problem means that most
of supervised domain adaptation methods are inapplicable. Our projection
domain shift problem differs from the conventional domain shift problems
in that (i) it is indirectly observed in terms of the projection
shift rather than the feature distribution shift, and (ii) the source
domain classes and target domain classes are completely different
and could even be unrelated. Consequently our domain adaptation method
differs significantly from the existing unsupervised ones such as
\cite{fernando2013unsupDAsubspace} in that our method relies on correlating different representations of the unlabelled
target data in a multi-view embedding space. %In the context of natural language processing, Blitzer et al.~\cite{blitzer2009zsDA} studied predicting user ratings in different domains (e.g.~book and DVD reviews). In their work views correspond to domains whilst our views correspond to representations. In addition, their tasks for  the source and target domains are the same -- estimating user ratings, whilst we aim to recognise a different set of object classes in the target domain.

\noindent \textbf{Learning multi-view embedding spaces}\quad{}Relating
low-level feature and semantic views of data has been exploited in
visual recognition and cross-modal retrieval. Most existing work \cite{SocherFeiFeiCVPR2010,multiviewCCAIJCV,HwangIJCV,topicimgannot}
focuses on modelling images/videos with associated text (e.g. tags
on Flickr/YouTube). Multi-view CCA is often exploited to provide unsupervised
fusion of different modalities. However, there are two fundamental
differences between previous multi-view embedding work and ours: (1)
Our embedding space is transductive, that is, learned from unlabelled
target data from which all semantic views are estimated by projection
rather than being the original views. These projected views thus have
the projection domain shift problem that the previous work does not
have. (2) The objectives are different: we aim to rectify the projection
domain shift problem via the embedding in order to perform better
recognition and annotation while previous studies target primarily
cross-modal retrieval. Note that although in this work, the popular CCA model is adopted for multi-view embedding, other models \cite{Rosipal2006,DBLP:conf/iccv/WangHWWT13}
could also be considered.}

\noindent \textbf{Graph-based label propagation}\quad{}In most previous
zero-shot learning studies (e.g., direct attribute prediction (DAP)
\cite{lampert13AwAPAMI}), the available knowledge (a single 
prototype per target class) is very limited. There has therefore been
recent interest in additionally exploiting the unlabelled target data
distribution by transductive learning \cite{transferlearningNIPS,yanweiPAMIlatentattrib}.
However, both \cite{transferlearningNIPS} and \cite{yanweiPAMIlatentattrib}
suffer from the projection domain shift problem, and are unable to
effectively exploit multiple semantic representations/views. In contrast, after
embedding, our framework synergistically
integrates the low-level feature and semantic representations by transductive
multi-view hypergraph label propagation (TMV-HLP). Moreover, TMV-HLP
generalises beyond zero-shot to N-shot learning if labelled instances
are available for the target classes.

In a broader context, graph-based label propagation \cite{zhou2004graphLabelProp}
in general, and classification on multi-view graphs (C-MG) in particular
 are well-studied  in semi-supervised learning. Most
C-MG solutions are based on the seminal work of Zhou \emph{et al}
\cite{Zhou2007ICML} which generalises spectral clustering from a
single graph to multiple graphs by defining a mixture of random walks
on multiple graphs. In the embedding space, instead
of constructing local neighbourhood graphs for each view independently
(e.g.~TMV-BLP \cite{embedding2014ECCV}), this paper proposes a {\emph{distributed
representation} of pairwise similarity using heterogeneous
hypergraphs. Such a distributed heterogeneous hypergraph representation
can better explore the higher-order relations between any two nodes
of different complementary views, and thus give rise to a more robust pairwise similarity
graph and lead to better classification performance than previous
multi-view graph methods \cite{Zhou2007ICML,embedding2014ECCV}.
Hypergraphs have been used as an effective tool to align multiple
data/feature modalities in data mining \cite{Li2013a}, multimedia
\cite{fu2010summarize} and computer vision \cite{DBLP:journals/corr/LiLSDH13,Hong:2013:MHL:2503901.2503960}
applications. A hypergraph is the generalisation of a 2-graph with
edges connecting many nodes/vertices, versus connecting two nodes
in conventional 2-graphs. This makes it cope better with noisy nodes
and thus achieve better performance than conventional graphs \cite{videoObjHypergraph,ImgRetrHypergraph,fu2010summarize}.
The only existing work considering hypergraphs for multi-view data
modelling is \cite{Hong:2013:MHL:2503901.2503960}. Different from
the multi-view hypergraphs proposed in \cite{Hong:2013:MHL:2503901.2503960}
which are homogeneous, that is, constructed in each view independently,
we construct a multi-view heterogeneous hypergraph: using the nodes
from one view as query nodes to compute hyperedges in another view.
This novel graph structure better exploits the complementarity of
different views in the common embedding space.

\section{Learning a Transductive Multi-View Embedding Space}
A schematic overview of our framework is given in
Fig.~\ref{fig:The-pipeline-of}. We next introduce some notation
and assumptions, followed by the details of how to map image features
into each semantic space, and how to map multiple spaces into a common
embedding space.

\subsection{Problem setup \label{sub:Problem-setup}}

We have $c_{S}$ source/auxiliary classes with $n_{S}$ instances
$S=\{X_{S},Y_{S}^{i},\mathbf{z}_{S}\}$ and $c_{T}$ target classes
$T=\left\{ X_{T},Y_{T}^{i},\mathbf{z}_{T}\right\} $ with $n_{T}$
instances.  $X_{S} \in \Bbb{R}^{n_{s}\times t}$ and $X_{T}\in \Bbb{R}^{n_{T}\times t}$ denote the $t-$dimensional low-level feature vectors of auxiliary and target instances respectively.
$\mathbf{z}_{S}$ and $\mathbf{z}_{T}$ are the auxiliary and target
class label vectors. We assume the auxiliary and target classes are
disjoint: $\mathbf{z}_{S}\cap\mathbf{z}_{T}=\varnothing$. We have
$I$ different types of  semantic representations; $Y_{S}^{i}$
and $Y_{T}^{i}$ represent the $i$-th type of $m_{i}$-dimensional
semantic representation for the auxiliary and target datasets respectively;
so $Y_{S}^{i}\in \Bbb{R}^{n_{S}\times m_{i}}$ and $Y_{T}^{i}\in \Bbb{R}^{n_{T}\times m_{i}}$.
Note that for the auxiliary dataset, $Y_{S}^{i}$ is given as each
data point is labelled. But for the target dataset, $Y_{T}^{i}$ is
missing, and its prediction $\hat{Y}_{T}^{i}$ from $X_{T}$ is used
instead. As we shall see, this is obtained using a projection
function learned from the auxiliary dataset. The problem of zero-shot
learning is to estimate $\mathbf{z}_{T}$ given $X_{T}$ and $\hat{Y}_{T}^{i}$.

Without any labelled data for the target classes, external knowledge
is needed to represent what each target class looks like, in the form
of class prototypes. Specifically, each target class $c$ has a pre-defined
class-level semantic prototype $\mathbf{y}_{c}^{i}$ in each semantic
view $i$. In this paper, we consider two types of intermediate semantic
representation (i.e.~$I=2$) -- attributes and word vectors, which
represent two distinct and complementary sources of information. We
use $\mathcal{X}$, $\mathcal{A}$ and $\mathcal{V}$ to denote the
low-level feature, attribute and word vector spaces respectively.
The attribute space $\mathcal{A}$ is typically manually defined using
a standard ontology. For the word vector space $\mathcal{V}$, we
employ the state-of-the-art skip-gram neural network model \cite{wordvectorICLR}
trained on all English Wikipedia articles%
\footnote{To 13 Feb. 2014, it includes 2.9 billion words from a 4.33 million-words
vocabulary (single and bi/tri-gram words).%
}. Using this learned model, we can project the textual name of any
class into the $\mathcal{V}$ space to get its word vector representation.
Unlike semantic attributes, it is a `free' semantic representation
in that this process does not need any human annotation. We next address
how to project low-level features into these two spaces.

\begin{figure*}
\begin{centering}
\includegraphics[scale=0.45]{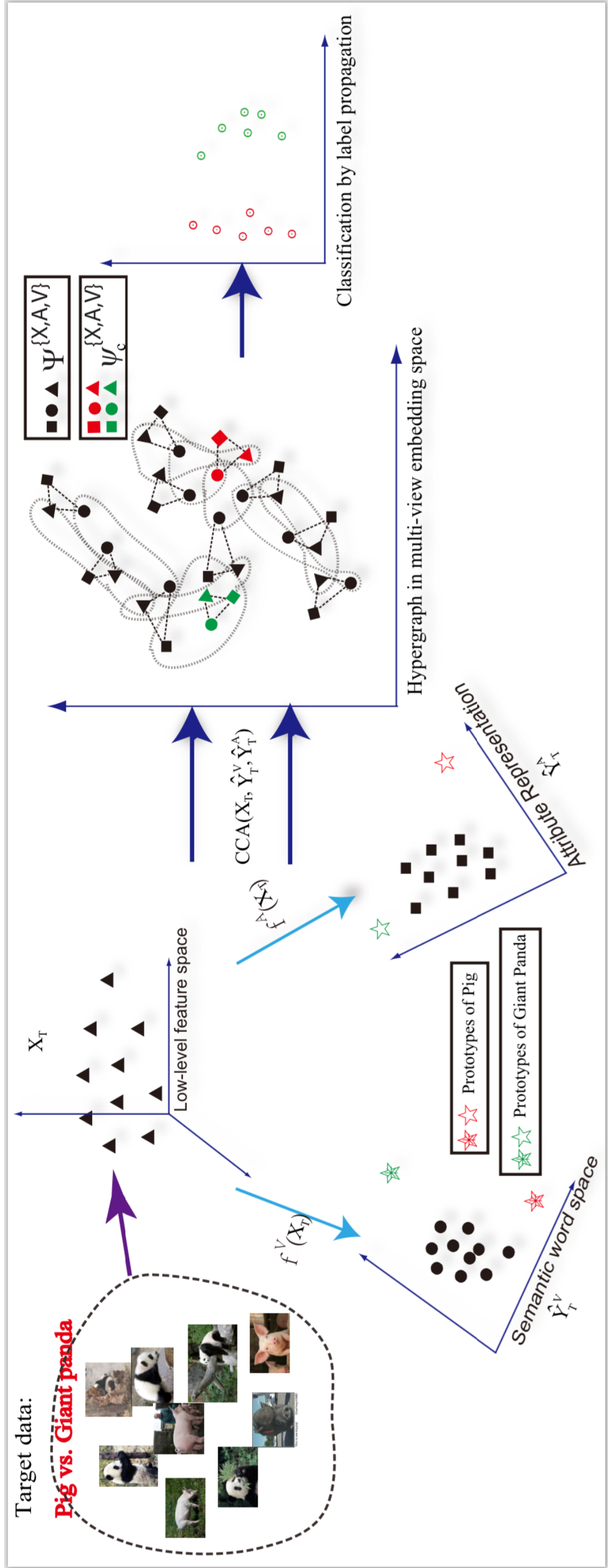}
\par\end{centering}

\caption{\label{fig:The-pipeline-of}The pipeline of our framework illustrated on the task of classifying  unlabelled target data into two classes.}
\end{figure*}

\subsection{Learning the projections of semantic spaces }

Mapping images and videos into semantic space $i$ requires a projection
function $f^{i}:\mathcal{X}\to\mathcal{Y}^{i}$. This is typically
realised by classifier \cite{lampert2009zeroshot_dat} or regressor
\cite{RichardNIPS13}. In this paper, using the auxiliary set $S$,
we train support vector classifiers $f^{\mathcal{A}}(\cdot)$ and
support vector regressors $f^{\mathcal{V}}(\cdot)$ for each dimension\footnote{Note that methods for learning projection functions for all dimensions
jointly exist (e.g.~\cite{DeviseNIPS13}) and can be adopted in our
framework.} of the auxiliary class attribute and word vectors respectively. Then the target class instances $X_{T}$ have the semantic projections:
$\hat{Y}_{T}^{\mathcal{A}}=f^{\mathcal{A}}(X_{T})$ and $\hat{Y}_{T}^{\mathcal{V}}=f^{\mathcal{V}}(X_{T})$.
However, these predicted intermediate semantics have the projection
domain shift problem illustrated in Fig.~\ref{fig:domain-shift:Low-level-feature-distribution}.
To address this, we learn a transductive multi-view semantic embedding
space to align the semantic projections with the low-level features
of target data.

\subsection{Transductive multi-view embedding }

We introduce a multi-view semantic alignment (i.e.
transductive multi-view embedding) process to correlate target instances
in different (biased) semantic view projections with their low-level
feature view. This process alleviates the projection domain shift
problem, as well as providing a common space in which heterogeneous
views can be directly compared, and their complementarity exploited
(Sec.~\ref{sec:Recognition-by-Multi-view}). To this end, we employ multi-view
Canonical Correlation Analysis (CCA) for $n_{V}$ views, with the
target data representation in view $i$ denoted ${\Phi}^{i}$, a $n_{T}\times m_{i}$
matrix. Specifically, we project three views of each
target class instance $f^{\mathcal{A}}(X_{T})$, $f^{\mathcal{V}}(X_{T})$
and $X_{T}$ (i.e.~$n_{V}=I+1=3$) into a shared embedding space.
The three projection functions $W^{i}$ are learned by 
\begin{eqnarray}
\mathrm{\underset{\left\{ W^{i}\right\} _{i=1}^{n_{V}}}{min}} & \sum_{i,j=1}^{n_{V}} & Trace(W^{i}\Sigma_{ij}W^{j})\nonumber \\
= & \sum_{i,j=1}^{n_{V}} & \parallel{\Phi}^{i}W^{i}-{\Phi}^{j}W^{j}\parallel_{F}^{2}\nonumber \\
\mathrm{s.t.} & \left[W^{i}\right]^{T}\Sigma_{ii}W^{i}=I & \left[\mathbf{w}_{k}^{i}\right]^{T}\Sigma_{ij}\mathbf{w}_{l}^{j}=0\nonumber \\
i\neq j,k\neq l & i,j=1,\cdots,n_{V} & k,l=1,\cdots,n_{T}\label{eq:multi-viewCCA}
\end{eqnarray}
where $W^{i}$ is the projection matrix which maps the view ${\Phi}^{i}$
($\in \Bbb{R}^{n_{T}\times m_{i}}$) into the embedding space and $\mathbf{w}_{k}^{i}$
is the $k$th column of $W^{i}$\textcolor{black}{.} $\Sigma_{ij}$
is the covariance matrix between ${\Phi}^{i}$ and ${\Phi}^{j}$.
The optimisation problem above is multi-convex as long as $\Sigma_{ii}$
are non-singular. The local optimum can be easily found by iteratively
maximising over each $W^{i}$ given the current values of the other
coefficients as detailed in \cite{CCAoverview}. 

The dimensionality $m_{e}$ of the embedding space
is the sum of the input view dimensions, i.e.~$m_{e}=\sum_{i=1}^{n_{V}}m_{i}$,
so $W^{i}\in \Bbb{R}^{m_{i}\times m_{e}}$. Compared to the classic approach
to CCA \cite{CCAoverview} which projects to a lower dimension space,
this retains all the input information including uncorrelated dimensions
which may be valuable and complementary. Side-stepping the task of
explicitly selecting a subspace dimension, we use a more stable and
effective soft-weighting strategy to implicitly emphasise significant
dimensions in the embedding space. This can be seen as a generalisation
of standard dimension reducing approaches to CCA, which implicitly
define a binary weight vector that activates a subset of dimensions
and deactivates others. Since the importance of each dimension is
reflected by its corresponding eigenvalue \cite{CCAoverview,multiviewCCAIJCV},
we use the eigenvalues to weight the dimensions and define a \emph{weighted
embedding space} $\Gamma$: 
\begin{equation}
{\Psi}^{i}={\Phi}^{i}W^{i}\left[D^{i}\right]^{\lambda}={\Phi}^{i}W^{i}\tilde{D}^{i},\label{eq:ccamapping}
\end{equation}
where $D^{i}$ is a diagonal matrix with its diagonal elements set
to the eigenvalues of each dimension in the embedding space, $\lambda$
is a power weight of $D^{i}$ and empirically set to $4$ \cite{multiviewCCAIJCV},
and ${\Psi}^{i}$ is the final representation of the target data from
view $i$ in $\Gamma$. We index the $n_{V}=3$ views as $i\in\{\mathcal{X},\mathcal{V},\mathcal{A}\}$
for notational convenience. The same formulation can be used if more
views are available.

\noindent \textbf{Similarity in the embedding space}\quad{}The choice
of similarity metric is important for high-dimensional embedding spaces.  For the subsequent recognition and annotation
tasks, we compute cosine distance in $\Gamma$ by $l_{2}$ normalisation:
normalising any vector $\bm{\psi}_{k}^{i}$ (the $k$-th row of ${\Psi}^{i}$) to unit length (i.e.~$\parallel\bm{\psi}_{k}^{i}\parallel_{2}=1$).
Cosine similarity is given by the inner product of any two vectors
in $\Gamma$.

\section{Recognition by Multi-view Hypergraph Label Propagation \label{sec:Recognition-by-Multi-view}}

For zero-shot recognition, each target class $c$ to be recognised
has a semantic prototype $\mathbf{y}_{c}^{i}$ in each view $i$.
Similarly, we have three views of each unlabelled instance $f^{\mathcal{A}}(X_{T})$,
$f^{\mathcal{V}}(X_{T})$ and $X_{T}$. The class prototypes are expected
to be the mean of the distribution of their class in semantic space,
since the projection function $f^{i}$ is trained to map instances
to their class prototype in each semantic view. To exploit the learned
space $\Gamma$ to improve recognition, we project both the unlabelled
instances and the prototypes into the embedding space\textcolor{red}{}%
\footnote{Before being projected into $\Gamma$, the prototypes
are updated by semi-latent zero shot learning algorithm in~\cite{yanweiPAMIlatentattrib}.}%
}. The prototypes $\mathbf{y}_{c}^{i}$ for views $i\in\{\mathcal{A},\mathcal{V}\}$
are projected as $\bm{\psi}_{c}^{i}=\mathbf{y}_{c}^{i}W^{i}\tilde{D}^{i}$.
So we have $\bm{\psi}_{c}^{\mathcal{A}}$ and $\bm{\psi}_{c}^{\mathcal{\mathcal{V}}}$
for the attribute and word vector prototypes of each target class
$c$ in $\Gamma$. In the absence of a prototype for the (non-semantic)
low-level feature view $\mathcal{X}$, we synthesise it as $\bm{\psi}_{c}^{\mathcal{X}}=(\bm{\psi}_{c}^{\mathcal{A}}+\bm{\psi}_{c}^{\mathcal{\mathcal{V}}})/2$.
If labelled data is available (i.e., N-shot case), these are also projected
into the space. Recognition could now be achieved using NN classification
with the embedded prototypes/N-shots as labelled data. However, this
does not effectively exploit the multi-view complementarity, and suffers
from labelled data (prototype) sparsity. To solve this problem, we next introduce a unified
framework to fuse the views and transductively exploit the manifold
structure of the unlabelled target data to perform zero-shot and N-shot
learning.

Most or all of the target instances are unlabelled, so classification
based on the sparse prototypes is effectively a semi-supervised learning
problem \cite{zhu2007sslsurvey}. We leverage graph-based semi-supervised
learning to exploit the manifold structure of the unlabelled data
transductively for classification. This differs from the conventional
approaches such as direct attribute prediction (DAP) \cite{lampert13AwAPAMI}
or NN, which too simplistically assume that the data distribution
for each target class is Gaussian or multinomial.  However, since
our embedding space contains multiple projections of the target data
and prototypes, it is hard to define a single graph that synergistically
exploits the manifold structure of all views. We therefore construct
multiple graphs within and across views in a transductive multi-view
hypergraph label propagation (TMV-HLP) model. Specifically, we construct the
heterogeneous hypergraphs across views to combine/align the different
manifold structures so as to enhance the robustness and exploit the
complementarity of different views. Semi-supervised learning is then
performed by propagating the labels from the sparse prototypes
(zero-shot) and/or the few labelled target instances (N-shot) to the
unlabelled data using random walk on the graphs.

\begin{figure*}[t]
\centering{}\includegraphics[scale=0.4]{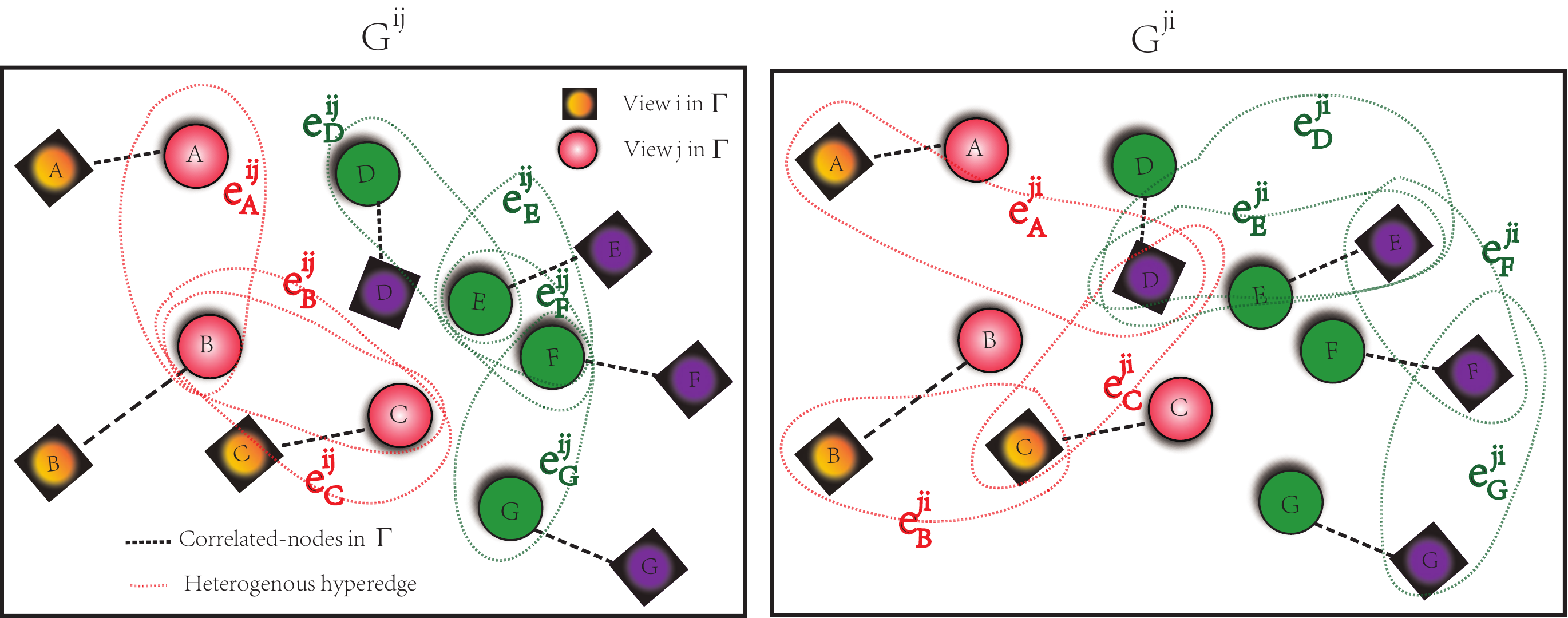}
\caption{\label{fig:Outliers-illustrations}An example of constructing heterogeneous
hypergraphs. Suppose in the embedding space, we have 14 nodes belonging
to 7 data points $A$, $B$, $C$, $D$, $E$, $F$ and $G$ of two
views -- view $i$ (rectangle) and view $j$ (circle). Data points
$A$,$B$,$C$ and $D$,$E$,$F$,$G$ belong to two different classes
-- red and green respectively. The multi-view semantic embedding maximises
the correlations (connected by black dash lines) between the two views
of the same node. Two hypergraphs are shown ($\mathcal{G}^{ij}$ at
the left and $\mathcal{G}^{ji}$ at the right) with the heterogeneous
hyperedges drawn with red/green dash ovals for the nodes of red/green
classes. Each hyperedge consists of two most similar nodes to the
query node. }
\end{figure*}

\subsection{Constructing heterogeneous hypergraphs}

\label{sub:Heterogenous-sub-hypergraph} \textbf{Pairwise node similarity}\quad{}The
key idea behind a hypergraph based method is to group similar data
points, represented as vertices/nodes on a graph, into hyperedges, so that the subsequent computation is less sensitive to individual noisy nodes. 
With the hyperedges, the pairwise similarity between two data points
are measured as the similarity between the two hyperedges that they
belong to, instead of that between the two nodes only. For both forming
hyperedges and computing the similarity between two hyperedges, pairwise
similarity between two graph nodes needs to be defined. In our embedding
space $\Gamma$, each data point in each view defines a node, and
the similarity between any pair of nodes is: 
\begin{equation}
\omega(\bm{\psi}_{k}^{i},\bm{\psi}_{l}^{j})=\exp(\frac{<\bm{\psi}_{k}^{i},\bm{\psi}_{l}^{j}>^{2}}{\varpi})\label{eq:sim_graph}
\end{equation}
where $<\bm{\psi}_{k}^{i},\bm{\psi}_{l}^{j}>^{2}$ is the square of
inner product between the $i$th and $j$th projections of nodes $k$
and $l$ with a bandwidth parameter $\varpi$%
\footnote{Most previous work \cite{transferlearningNIPS,Zhou2007ICML} sets
$\varpi$ by cross-validation. Inspired by \cite{lampertTutorial},
a simpler strategy for setting $\varpi$ is adopted: $\varpi\thickapprox\underset{k,l=1,\cdots,n}{\mathrm{median}}<\bm{\psi}_{k}^{i},\bm{\psi}_{l}^{j}>^{2}$
in order to have roughly the same number of similar and dissimilar
sample pairs. This makes the edge weights from different pairs of
nodes more comparable.%
}. Note that Eq (\ref{eq:sim_graph}) defines the pairwise similarity
between any two nodes within the same view ($i=j$) or across different
views ($i\neq j$).

\noindent \textbf{Heterogeneous hyperedges}\quad{}Given the multi-view
projections of the target data, we aim to construct a set of across-view
heterogeneous hypergraphs 
\begin{equation}
\mathcal{G}^{c}=\left\{ \mathcal{G}^{ij}\mid i,j\in\left\{ \mathcal{X},\mathcal{V},\mathcal{A}\right\} ,i\neq j\right\} \label{eq:hypergraph_def}
\end{equation}
 where $\mathcal{G}^{ij}=\left\{ \Psi^{i},E^{ij},\Omega^{ij}\right\} $
denotes the cross-view heterogeneous hypergraph from view $i$ to
$j$ (in that order) and $\Psi^{i}$ is the node
set in view $i$; $E^{ij}$ is the hyperedge set and $\Omega^{ij}$
is the pairwise node similarity set for the hyperedges. Specifically,
we have the hyperedge set $E^{ij}=\left\{ e_{\bm{\psi}_{k}^{i}}^{ij}\mid i\neq j,\: k=1,\cdots n_{T}+c_{T}\right\} $
where each hyperedge $e_{\bm{\psi}_{k}^{i}}^{ij}$ includes the nodes%
\footnote{Both the unlabelled samples and the prototypes are
nodes.}  in view $j$ that are the most similar to node $\bm{\psi}_{k}^{i}$
in view $i$ and the similarity set $\Omega^{ij}=\left\{ \Delta_{\bm{\psi}_{k}^{i}}^{ij}=\left\{ \omega\left(\bm{\psi}_{k}^{i},\bm{\psi}_{l}^{j}\right)\right\} \mid i\neq j,\bm{\psi}_{l}^{j}\in e_{\bm{\psi}_{k}^{i}}^{ij}\: k=1,\cdots n_{T}+c_{T}\right\} $
where $\omega\left(\bm{\psi}_{k}^{i},\bm{\psi}_{l}^{j}\right)$ is
computed using Eq (\ref{eq:sim_graph}).

\noindent  We call $\bm{\psi}_{k}^{i}$ the query
node for hyperedge $e_{\bm{\psi}_{k}^{i}}^{ij}$, since the hyperedge
$e_{\bm{\psi}_{k}^{i}}^{ij}$ intrinsically groups all nodes in view
$j$ that are most similar to node $\bm{\psi}_{k}^{i}$ in view $i$.
Similarly, $\mathcal{G}^{ji}$ can be constructed by using nodes
from view $j$ to query nodes in view $i$. Therefore given three
views, we have six across view/heterogeneous hypergraphs. Figure \ref{fig:Outliers-illustrations}
illustrates two heterogeneous hypergrahs constructed from two views.
Interestingly, our way of defining hyperedges naturally corresponds
to the star expansion \cite{hypergraphspectral} where the query node
(i.e.~$\bm{\psi}_{k}^{i}$) is introduced to connect each node in
the hyperedge $e_{\bm{\psi}_{k}^{i}}^{ij}$.

\noindent \textbf{Similarity strength of hyperedge}\quad{}For each hyperedge
$e_{\bm{\psi}_{k}^{i}}^{ij}$, we measure its similarity strength
by using its query nodes $\bm{\psi}_{k}^{i}$.  Specifically,
we use the weight $\delta_{\bm{\psi}_{k}^{i}}^{ij}$ to indicate the
similarity strength of nodes connected within the hyperedge $e_{\bm{\psi}_{k}^{i}}^{ij}$.
Thus, we define $\delta_{\bm{\psi}_{k}^{i}}^{ij}$ based on the mean
similarity of the set $\Delta_{\bm{\psi}_{k}^{i}}^{ij}$ for the hyperedge
\begin{align}
\delta_{\bm{\psi}_{k}^{i}}^{ij} & =\frac{1}{\mid e_{\bm{\psi}_{k}^{i}}^{ij}\mid}\sum_{\omega\left(\bm{\psi}_{k}^{i},\bm{\psi}_{l}^{j}\right)\in\Delta_{\bm{\psi}_{k}^{i}}^{ij},\bm{\psi}_{l}^{j}\in e_{\bm{\psi}_{k}^{i}}^{ij}}\omega\left(\bm{\psi}_{k}^{i},\bm{\psi}_{l}^{j}\right),\label{eq:heterogenous_similarity_weight}
\end{align}
where $\mid e_{\bm{\psi}_{k}^{i}}^{ij}\mid$ is the cardinality of
hyperedge $e_{\bm{\psi}_{k}^{i}}^{ij}$.

\noindent In the embedding space $\Gamma$, similarity
sets $\Delta_{\bm{\psi}_{k}^{i}}^{ij}$ and $\Delta_{\bm{\psi}_{l}^{i}}^{ij}$
can be compared. Nevertheless, these sets come from heterogeneous views
and have varying scales. Thus some normalisation steps are necessary
to make the two similarity sets more comparable and the subsequent
computation more robust. Specifically, we extend zero-score normalisation
to the similarity sets: (a) We assume $\forall\Delta_{\bm{\psi}_{k}^{i}}^{ij}\in\Omega^{ij}$
and $\Delta_{\bm{\psi}_{k}^{i}}^{ij}$ should follow Gaussian distribution.
Thus, we enforce zero-score normalisation to $\Delta_{\bm{\psi}_{k}^{i}}^{ij}$.
(b) We further assume that the retrieved similarity set $\Omega^{ij}$
between all the queried nodes $\bm{\psi}_{k}^{i}$ ($l=1,\cdots n_{T})$
from view $i$ and $\bm{\psi}_{l}^{j}$ should also follow Gaussian
distributions. So we again enforce Gaussian distribution to the pairwise
similarities between $\bm{\psi}_{l}^{j}$ and all query nodes from
view $i$ by zero-score normalisation. (c) We select the first $K$
highest values from $\Delta_{\bm{\psi_{k}^{i}}}^{ij}$ as new similarity
set $\bar{\Delta}_{\bm{\psi}_{k}^{i}}^{ij}$ for hyperedge $e_{\bm{\psi}_{k}^{i}}^{ij}$.
$\bar{\Delta}_{\bm{\psi}_{k}^{i}}^{ij}$ is then used in Eq (\ref{eq:heterogenous_similarity_weight})
in place of ${\Delta}_{\bm{\psi}_{k}^{i}}^{ij}$. These normalisation
steps aim to compute a more robust similarity between each pair of
hyperedges.

\noindent \textbf{Computing similarity between
hyperedges} \quad{} With the hypergraph, the similarity between two nodes is computed
 using their hyperedges $e_{\bm{\psi}_{k}^{i}}^{ij}$.	
Specifically, for each hyperedge there is an associated incidence
matrix $H^{ij}=\left(h\left(\bm{\psi}_{l}^{j},e_{\bm{\psi}_{k}^{i}}^{ij}\right)\right)_{(n_{T}+c_{T})\times\mid E^{ij}\mid}$
where 
\begin{equation}
h\left(\bm{\psi}_{l}^{j},e_{\bm{\psi}_{k}^{i}}^{ij}\right)=\begin{cases}
\begin{array}[t]{cc}
1 & if\:\bm{\psi}_{l}^{j}\in e_{\bm{\psi}_{k}^{i}}^{ij}\\
0 & otherwise
\end{array}\end{cases}\label{eq:heterogenous_hard_incidence_matrix}
\end{equation}
To take into consideration the similarity strength between hyperedge
and query node, we extend the binary valued hyperedge incidence matrix
$H^{ij}$ to soft-assigned incidence matrix $SH^{ij}=\left(sh\left(\bm{\psi}_{l}^{j},e_{\bm{\psi}_{k}^{i}}^{ij}\right)\right)_{(n_{T}+c_{T})\times\mid E^{ij}\mid}$
as follows 
\begin{equation}
sh\left(\bm{\psi}_{l}^{j},e_{\bm{\psi}_{k}^{i}}^{ij}\right)=\delta_{\bm{\psi}_{k}^{i}}^{ij}\cdot\omega\left(\bm{\psi}_{k}^{i},\bm{\psi}_{l}^{j}\right)\cdot h\left(\bm{\psi}_{l}^{j},e_{\bm{\psi}_{k}^{i}}^{ij}\right)\label{eq:soft_incident_matrix}
\end{equation}
This soft-assigned incidence matrix is the product of three components:
(1) the weight $\delta_{\bm{\psi}_{k}^{i}}$ for hyperedge $e_{\bm{\psi}_{k}^{i}}^{ij}$;
(2) the pairwise similarity computed using queried node $\bm{\psi}_{k}^{i}$;
(3) the binary valued hyperedge incidence matrix element $h\left(\bm{\psi}_{l}^{j},e_{\bm{\psi}_{k}^{i}}^{ij}\right)$.
To make the values of $SH^{ij}$ comparable among the different heterogeneous
views, we apply $l_{2}$ normalisation to the soft-assigned incidence
matrix values for all node incident to each hyperedge.

Now for each heterogeneous hypergraph, we can finally define the pairwise
similarity between any two nodes or hyperedges. Specifically for $\mathcal{G}^{ij}$,
the similarity between the $o$-th and $l$-th nodes is 
\begin{equation}
\omega_{c}^{ij}\left(\bm{\psi}_{o}^{j},\bm{\psi}_{l}^{j}\right)=\sum_{e_{\bm{\psi}_{k}^{i}}^{ij}\in E^{ij}}sh\left(\bm{\psi}_{o}^{j},e_{\bm{\psi}_{k}^{i}}^{ij}\right)\cdot sh\left(\bm{\psi}_{l}^{j},e_{\bm{\psi}_{k}^{i}}^{ij}\right).\label{eq:hyperedge_weights}
\end{equation}

With this pairwise hyperedge similarity, the hypergraph definition
is now complete. Empirically, given a node, other nodes on the graph that have very low similarities will have very limited effects on its label.
Thus, to reduce computational cost, we only use the K-nearest-neighbour
(KNN)\footnote{$K=30$. It can be varied from $10\sim50$ with
little effect in our experiments.} nodes of each node~\cite{zhu2007sslsurvey} for the subsequent label propagation step.

\noindent \textbf{The advantages of heterogeneous hypergraphs}\quad{}We
argue that the pairwise similarity of heterogeneous hypergraphs is
a distributed representation \cite{Bengio:2009:LDA:1658423.1658424}. To explain it, we can use star
extension \cite{hypergraphspectral} to extend a hypergraph into a
2-graph. For each hyperedge $e_{\bm{\psi}_{k}^{i}}^{ij}$, the query
node $\bm{\psi}_{k}^{i}$ is used to compute the pairwise similarity
$\Delta_{\bm{\psi}_{k}^{i}}^{ij}$ of all the nodes in view $j$.
Each hyperedge can thus define a hyper-plane by categorising the nodes
in view $j$ into two groups: strong and weak similarity group regarding
to query node $\bm{\psi}_{k}^{i}$. In other words, the hyperedge
set $E^{ij}$ is multi-clustering with linearly separated regions
(by each hyperplane) per classes. Since the final pairwise similarity
in Eq (\ref{eq:hyperedge_weights}) can be represented by a set of
similarity weights computed by hyperedge, and such weights are not
mutually exclusive and are statistically independent, we consider
the heterogeneous hypergraph a distributed representation. The advantage
of having a distributed representation has been studied by Watts and
Strogatz~\cite{Watts-Colective-1998,Watts.2004} which shows that
such a representation gives rise to better convergence rates and better
clustering abilities. In contrast, the homogeneous hypergraphs adopted
by previous work \cite{ImgRetrHypergraph,fu2010summarize,Hong:2013:MHL:2503901.2503960}
does not have this property which makes them less robust against noise.
In addition, fusing different views in the early stage of graph construction
potentially can lead to better exploitation of the complementarity
of different views.  However,
it is worth pointing out that (1) The reason we can query nodes across
views to construct heterogeneous hypergraph is because we have projected
all views in the same embedding space in the first place. (2) Hypergraphs
typically gain robustness at the cost of losing discriminative power
-- it essentially blurs the boundary of different clusters/classes
by taking average over hyperedges. A typical solution is to
fuse hypergraphs with 2-graphs~\cite{fu2010summarize,Hong:2013:MHL:2503901.2503960,Li2013a},
which we adopt here as well.

\subsection{Label propagation by random walk}

Now we have two types of graphs: heterogeneous hypergraphs $\mathcal{G}^{c}=\left\{ \mathcal{G}^{ij}\right\} $
and 2-graphs\footnote{That is the K-nearest-neighbour graph of each view
in $\Gamma$ \cite{embedding2014ECCV}.}% 
$\mathcal{G}^{p}=\left\{ \mathcal{G}^{i}\right\} $.
Given three views ($n_{V}=3$), we thus have nine graphs in total
(six hypergraphs and three 2-graphs). To classify
the unlabelled nodes, we need to propagate label information from the prototype
nodes across the graph. Such semi-supervised label propagation \cite{Zhou2007ICML,zhu2007sslsurvey}
has a closed-form solution and is explained as a random walk. %({\bf need to explain what  are typical alternative / baseline method for the problem before motivate why we choose random walk})
A random walk requires pairwise transition probability
for nodes $k$ and $l$. We obtain this by aggregating the information
from all graphs $\mathcal{G}=\left\{ \mathcal{G}^{p};\mathcal{G}^{c}\right\} $,
\begin{align}
p\left(k\rightarrow l\right) & =\sum_{i\in\left\{ \mathcal{X},\mathcal{V},\mathcal{A}\right\} }p\left(k\rightarrow l\mid\mathcal{G}^{i}\right)\cdot p\left(\mathcal{G}^{i}\mid k\right)+\label{eq:transition_probability}\\
 & \sum_{i,j\in\left\{ \mathcal{X},\mathcal{V},\mathcal{A}\right\} ,i\ne j}p\left(k\rightarrow l\mid\mathcal{G}^{ij}\right)\cdot p\left(\mathcal{G}^{ij}\mid k\right)\nonumber 
\end{align}
where 
\begin{equation}
p\left(k\rightarrow l\mid\mathcal{G}^{i}\right)=\frac{\omega_{p}^{i}(\bm{\psi}_{k}^{i},\bm{\psi}_{l}^{i})}{\sum_{o}\omega_{p}^{i}(\bm{\psi}_{k}^{i},\bm{\psi}_{o}^{i})},\label{eq:prob_graphs-1}
\end{equation}
and
\[
p\left(k\rightarrow l\mid\mathcal{G}^{ij}\right)=\frac{\omega_{c}^{ij}(\bm{\psi}_{k}^{j},\bm{\psi}_{l}^{j})}{\sum_{o}\omega_{c}^{ij}(\bm{\psi}_{k}^{j},\bm{\psi}_{o}^{j})}
\]
and then the posterior probability to choose graph $\mathcal{G}^{i}$
at projection/node $\bm{\psi}_{k}^{i}$ will be: 
\begin{align}
p(\mathcal{G}^{i}|k) & =\frac{\pi(k|\mathcal{G}^{i})p(\mathcal{G}^{i})}{\sum_{i}\pi(k|\mathcal{G}^{i})p(\mathcal{G}^{i})+\sum_{ij}\pi(k|\mathcal{G}^{ij})p(\mathcal{G}^{ij})}\label{eq:post_prob_i}\\
p(\mathcal{G}^{ij}|k) & =\frac{\pi(k|\mathcal{G}^{ij})p(\mathcal{G}^{ij})}{\sum_{i}\pi(k|\mathcal{G}^{i})p(\mathcal{G}^{i})+\sum_{ij}\pi(k|\mathcal{G}^{ij})p(\mathcal{G}^{ij})}
\end{align}
\noindent where $p(\mathcal{G}^{i})$ and $p(\mathcal{G}^{ij})$ are
the prior probability of graphs $\mathcal{G}^{i}$ and $\mathcal{G}^{ij}$
in the random walk. This probability expresses prior expectation about
the informativeness of each graph. The same\emph{ }Bayesian model
averaging \cite{embedding2014ECCV} can be used here to estimate these
prior probabilities. However, the computational cost is combinatorially
increased with the number of views; and it turns out the prior is
not critical to the results of our framework. Therefore, uniform prior
is used in our experiments.

The stationary probabilities for node $k$ in $\mathcal{G}^{i}$ and
$\mathcal{G}^{ij}$ are 
\begin{align}
\pi(k|\mathcal{G}^{i}) & =\frac{\sum_{l}\omega_{p}^{i}(\bm{\psi}_{k}^{i},\bm{\psi}_{l}^{i})}{\sum_{o}\sum_{l}\omega_{p}^{i}(\bm{\psi}_{k}^{i},\bm{\psi}_{o}^{i})}\label{eq:stati_prob_i}\\
\pi(k|\mathcal{G}^{ij}) & =\frac{\sum_{l}\omega_{c}^{ij}(\bm{\psi}_{k}^{j},\bm{\psi}_{l}^{j})}{\sum_{k}\sum_{o}\omega_{c}^{ij}(\bm{\psi}_{k}^{j},\bm{\psi}_{o}^{j})}\label{eq:heterogeneous_stationary_prob}
\end{align}

Finally, the stationary probability across the multi-view hypergraph
is computed as: 
\begin{align}
\pi(k) & =\sum_{i\in\left\{ \mathcal{X},\mathcal{V},\mathcal{A}\right\} }\pi(k|\mathcal{G}^{i})\cdot p(\mathcal{G}^{i})+\label{eq:stat_prob}\\
 & \sum_{i,j\in\left\{ \mathcal{X},\mathcal{V},\mathcal{A}\right\} ,i\neq j}\pi(k|\mathcal{G}^{ij})\cdot p(\mathcal{G}^{ij})
\end{align}

\noindent Given the defined graphs and random walk process, we can
derive our label propagation algorithm (TMV-HLP). Let $P$ denote
the transition probability matrix defined by Eq (\ref{eq:transition_probability})
and $\Pi$ the diagonal matrix with the elements $\pi(k)$ computed
by Eq (\ref{eq:stat_prob}). The Laplacian matrix $\mathcal{L}$ combines
information of different views and is defined as: $\mathcal{L}=\Pi-\frac{\Pi P+P^{T}\Pi}{2}.$
The label matrix $Z$ for labelled N-shot data or zero-shot prototypes
is defined as: 
\begin{equation}
Z(q_{k},c)=\begin{cases}
\begin{array}{c}
1\\
-1\\
0
\end{array} & \begin{array}{c}
q_{k}\in class\, c\\
q_{k}\notin class\, c\\
unknown
\end{array}\end{cases}\label{eq:initial_label}
\end{equation}
Given the label matrix $Z$ and Laplacian $\mathcal{L}$, label propagation
on multiple graphs has the closed-form solution \cite{Zhou2007ICML}: $\hat{Z}=\eta(\eta\Pi+\mathcal{L})^{-1}\Pi Z$ where $\eta$ is
a regularisation parameter%
\footnote{It can be varied from $1-10$ with little effects in our experiments.}. Note that in our framework, both labelled target class instances
and prototypes are modelled as graph nodes. Thus the difference between
zero-shot and N-shot learning lies only on the initial labelled instances:
Zero-shot learning has the prototypes as labelled nodes; N-shot has
instances as labelled nodes; and a new condition exploiting both prototypes
and N-shot together is possible. This unified recognition framework
thus applies when either or both of prototypes and labelled instances
are available. The computational cost of our TMV-HLP
is $\mathcal{O}\left((c_{T}+n_{T})^{2}\cdot n_{V}^{2}+(c_{T}+n_{T})^{3}\right)$,
where $K$ is the number of nearest neighbours in the KNN graphs,
and $n_{V}$ is the number of views. It costs $\mathcal{O}((c_{T}+n_{T})^{2}\cdot n_{V}^{2})$
to construct the heterogeneous graph, while the inverse matrix of Laplacian
matrix $\mathcal{L}$ in label propagation step will take $\mathcal{O}((c_{T}+n_{T})^{3})$
computational time, which however can be further reduced to $\mathcal{O}(c_{T}n_{T}t)$ using the recent
work of Fujiwara et al.~\cite{FujiwaraICML2014efficientLP}, where $t$ is an iteration parameter
in their paper and $t\ll n_{T}$.

\section{Annotation and Beyond\label{sec:Annotation-and-Beyond}}

Our multi-view embedding space $\Gamma$ bridges the semantic gap
between low-level features $\mathcal{X}$ and semantic representations
$\mathcal{A}$ and $\mathcal{V}$. Leveraging this cross-view mapping,
annotation \cite{hospedales2011video_tags,topicimgannot,multiviewCCAIJCV}
can be improved and applied in novel ways. We consider three annotation
tasks here:

\noindent \textbf{Instance level annotation}\quad{}Given a new instance
$u$, we can describe/annotate it by predicting its attributes. The
conventional solution is directly applying $\hat{\mathbf{y}}_{u}^{\mathcal{A}}=f^{\mathcal{A}}(\mathbf{x}_{u})$
for test data $\mathbf{x}_{u}$ \cite{farhadi2009attrib_describe,multiviewCCAIJCV}.
However, as analysed before, this suffers from the projection domain
shift problem. To alleviate this, our multi-view embedding space aligns
the semantic attribute projections with the low-level features of
each unlabelled instance in the target domain. This alignment can
be used for image annotation in the target domain. Thus, with our
framework, we can now infer attributes for any test instance via the
learned embedding space $\Gamma$ as $\hat{\mathbf{y}}_{u}^{A}=\mathbf{x}_{u}W^{\mathcal{X}}\tilde{D}^{\mathcal{X}}\left[W^{\mathcal{A}}\tilde{D}^{\mathcal{A}}\right]^{-1}$.

\noindent \textbf{Zero-shot class description}\quad{}From a broader
machine intelligence perspective, one might be interested to ask what
are the attributes of an unseen class, based solely on the class name.
Given our multi-view embedding space, we
can infer the semantic attribute description of a novel class. This
\textit{zero-shot class description} task could be useful, for example,
to hypothesise the zero-shot attribute prototype of a class instead
of defining it by experts \cite{lampert2009zeroshot_dat} or ontology
\cite{yanweiPAMIlatentattrib}. Our transductive embedding enables
this task by connecting semantic word space (i.e.~naming) and discriminative
attribute space (i.e.~describing). Given the prototype $\mathbf{y}_{c}^{\mathcal{V}}$
from the name of a novel class $c$, we compute $\hat{\mathbf{y}}_{c}^{\mathcal{A}}=\mathbf{y}_{c}^{\mathcal{V}}W^{\mathcal{V}}\tilde{D}^{\mathcal{V}}\left[W^{\mathcal{A}}\tilde{D}^{\mathcal{A}}\right]^{-1}$
to generate the class-level attribute description.

\noindent \textbf{Zero prototype learning}\quad{}This task is the
inverse of the previous task -- to infer the name of class given a
set of attributes. It could be useful, for example, to validate or
assess a proposed zero-shot attribute prototype, or to provide an
automated semantic-property based index into a dictionary or database.
To our knowledge, this is the first attempt to evaluate the quality
of a class attribute prototype because no previous work has directly
and systematically linked linguistic knowledge space with visual attribute
space. Specifically given an attribute prototype $\mathbf{y}_{c}^{\mathcal{A}}$,
we can use $\hat{\mathbf{y}}_{c}^{\mathcal{V}}=\hat{\mathbf{y}}_{c}^{\mathcal{A}}W^{\mathcal{A}}\tilde{D}^{\mathcal{A}}\left[W^{\mathcal{V}}\tilde{D}^{\mathcal{V}}\right]^{-1}$
to name the corresponding class and perform retrieval on dictionary
words in $\mathcal{V}$ using $\hat{\mathbf{y}}_{c}^{\mathcal{V}}$.

\section{Experiments}

\subsection{Datasets and settings }

We evaluate our framework on three widely used image/video 
datasets: Animals with Attributes (AwA), Unstructured Social Activity
Attribute (USAA), and Caltech-UCSD-Birds (CUB). \textbf{AwA} \cite{lampert2009zeroshot_dat}
consists of $50$ classes of animals ($30,475$ images) and $85$ associated
class-level attributes. It has a standard source/target split for
zero-shot learning with $10$ classes and $6,180$ images held out
as the target dataset. We use the same `hand-crafted' low-level features
(RGB colour histograms, SIFT, rgSIFT, PHOG, SURF and local self-similarity
histograms) released with the dataset (denoted as $\mathcal{H}$);
and the same multi-kernel learning (MKL) attribute classifier from
\cite{lampert2009zeroshot_dat}.\textbf{ USAA} is a video dataset
\cite{yanweiPAMIlatentattrib} with $69$ instance-level attributes
for $8$ classes of complex (unstructured) social group activity videos
from YouTube. Each class has around $100$ training and test videos
respectively. USAA provides the instance-level attributes since there
are significant intra-class variations. We use the thresholded mean
of instances from each class to define a binary attribute prototype
as in \cite{yanweiPAMIlatentattrib}. The same setting in \cite{yanweiPAMIlatentattrib}
is adopted: $4$ classes as source and $4$ classes as target data.
We use exactly the same SIFT, MFCC and STIP low-level features for
USAA as in \cite{yanweiPAMIlatentattrib}. \textbf{CUB-200-2011} \cite{WahCUB_200_2011}
contains $11,788$ images of $200$ bird classes. This is more challenging
 than AwA -- it is designed for fine-grained recognition and
has more classes but fewer images. Each class is annotated with $312$
binary attributes derived from a bird species ontology. We use $150$
classes as auxiliary data, holding out $50$ as test data. We extract
$128$ dimensional SIFT and colour histogram descriptors from regular
grid of multi-scale and aggregate them into image-level feature Fisher
Vectors ($\mathcal{F}$) by using $256$ Gaussians, as in \cite{labelembeddingcvpr13}.
Colour histogram and PHOG features are also used to extract  global
color and texture cues from each image. Due to the recent progress
on deep learning based representations, we also extract  OverFeat
($\mathcal{O}$) \cite{sermanet-iclr-14}%
\footnote{We use the trained model of OverFeat in \cite{sermanet-iclr-14}.%
} from AwA and CUB as an alternative to $\mathcal{H}$ and $\mathcal{F}$
respectively. In addition,  DeCAF ($\mathcal{D}$) \cite{decaf}
 is also considered for AwA.

We report absolute classification accuracy on USAA and mean accuracy
for AwA and CUB for direct comparison to published results. The word
vector space is trained by the model in \cite{wordvectorICLR} with
$1,000$ dimensions.

\begin{table*}[ht]
\begin{centering}
\begin{tabular}{c|c|c|c|c|c|c}
\hline 
Approach  & \multicolumn{1}{c|}{AwA ($\mathcal{H}$ \cite{lampert2009zeroshot_dat})} & AwA ($\mathcal{O}$)  & AwA $\left(\mathcal{O},\mathcal{D}\right)$  & USAA  & CUB ($\mathcal{O}$)  & CUB ($\mathcal{F}$) \tabularnewline
\hline 
DAP  & 40.5(\cite{lampert2009zeroshot_dat}) / 41.4(\textcolor{black}{\cite{lampert13AwAPAMI})
/ 38.4{*}}  & 51.0{*}  & 57.1{*}  & 33.2(\cite{yanweiPAMIlatentattrib}) / 35.2{*}  & 26.2{*}  & 9.1{*}\tabularnewline
IAP  & 27.8(\cite{lampert2009zeroshot_dat}) / 42.2(\textcolor{black}{\cite{lampert13AwAPAMI})}  & --  & --  & --  & --  & --\tabularnewline
M2LATM \cite{yanweiPAMIlatentattrib} {*}{*}{*} & 41.3  & --  & --  & 41.9  & --  & --\tabularnewline
ALE/HLE/AHLE \cite{labelembeddingcvpr13}  & 37.4/39.0/43.5  & --  & --  & --  & --  & 18.0{*}\tabularnewline
Mo/Ma/O/D \cite{marcuswhathelps}  & 27.0 / 23.6 / 33.0 / 35.7  & --  & --  & --  & --  & --\tabularnewline
PST \cite{transferlearningNIPS} {*}{*}{*} & 42.7  & 54.1{*}  & 62.9{*}  & 36.2{*}  & 38.3{*}  & 13.2{*}\tabularnewline
\cite{Yucatergorylevel}  & 48.3{*}{*}  & --  & --  & --  & --  & --\tabularnewline
TMV-BLP \cite{embedding2014ECCV}{*}{*}{*} & 47.7 & 69.9 & 77.8 & 48.2 & 45.2 & 16.3\tabularnewline
\hline 
TMV-HLP {*}{*}{*} & \textbf{49.0}  & \textbf{73.5}  & \textbf{80.5}  & \textbf{50.4}  & \textbf{47.9}  & \textbf{19.5}\tabularnewline
\hline 
\end{tabular}
\par\end{centering}

\noindent \caption{\label{tab:Comparison-with-stateofart}Comparison with the state-of-the-art
on zero-shot learning on AwA, USAA and CUB. Features $\mathcal{H}$,
$\mathcal{O}$, $\mathcal{D}$ and $\mathcal{F}$ represent hand-crafted, OverFeat, DeCAF,
and Fisher Vector respectively. Mo, Ma, O and D represent the highest
results by the mined object class-attribute associations, mined attributes,
objectness as attributes and direct similarity methods used in \cite{marcuswhathelps}
respectively. `--': no result reported. {*}: our implementation. {*}{*}:
requires additional human annotations.{*}{*}{*}:
requires unlabelled data, i.e.~a transductive setting. }
\end{table*}

\subsection{Recognition by zero-shot learning }

\subsubsection{Comparisons with state-of-the-art}

We compare our method (TMV-HLP) with  the recent state-of-the-art
models that report results or can be re-implemented by us on the three
datasets in Table~\ref{tab:Comparison-with-stateofart}. They cover
a wide range of approaches on utilising semantic intermediate representation
for zero-shot learning. They can be roughly categorised according
to the semantic representation(s) used: DAP and IAP (\cite{lampert2009zeroshot_dat},
\cite{lampert13AwAPAMI}), M2LATM \cite{yanweiPAMIlatentattrib},
ALE \cite{labelembeddingcvpr13}, \cite{transferlearningNIPS} and
\cite{unifiedProbabICCV13} use attributes only; HLE/AHLE \cite{labelembeddingcvpr13}
and Mo/Ma/O/D \cite{marcuswhathelps} use both attributes and linguistic
knowledge bases (same as us); \cite{Yucatergorylevel} uses attribute
and some additional human manual annotation. Note that our linguistic
knowledge base representation is in the form of word vectors, which
does not incur additional manual annotation. Our method also does
not exploit data-driven attributes such as M2LATM \cite{yanweiPAMIlatentattrib}
and Mo/Ma/O/D \cite{marcuswhathelps}.

Consider first the results on the most widely used AwA. Apart
from the standard hand-crafted feature ($\mathcal{H}$),
we consider the more powerful OverFeat deep feature ($\mathcal{O}$),
and a combination of OverFeat and DeCAF $\left(\mathcal{O},\mathcal{D}\right)$%
\footnote{With these two low-level feature views, there are six views in total
in the embedding space.%
}. Table~\ref{tab:Comparison-with-stateofart} shows that (1) with
the same experimental settings and the same feature ($\mathcal{H}$),
our TMV-HLP ($49.0\%$) outperforms the best result reported so far
(48.3\%) in \cite{Yucatergorylevel} which, unlike ours, requires additional human
annotation to relabel the similarities between auxiliary and target
classes. 
(2) With the more powerful OverFeat feature, our method achieves $73.5\%$
zero-shot recognition accuracy. Even more remarkably, when both the
OverFeat and DeCAF features are used in our framework, the result
(see the AwA $\left(\mathcal{O},\mathcal{D}\right)$ column) is $80.5\%$.
Even with only 10 target classes, this is an extremely good result given
that we do not have any labelled samples from the target classes. Note
that this good result is not solely due to the feature strength, as
the margin between the conventional DAP and our TMV-HLP is much bigger
indicating that our TMV-HLP plays a critical role in achieving this
result. 
(3) Our method is also superior to the AHLE method in \cite{labelembeddingcvpr13}
which also uses two semantic spaces: attribute and WordNet hierarchy.
Different from our embedding framework, AHLE simply concatenates the
two spaces. (4) Our method also outperforms the other alternatives
of either mining other semantic knowledge bases (Mo/Ma/O/D \cite{marcuswhathelps})
or exploring data-driven attributes (M2LATM \cite{yanweiPAMIlatentattrib}).
(5) Among all compared methods, PST \cite{transferlearningNIPS} is
the only one except ours that performs label propagation based transductive
learning. It yields better results than DAP in all the experiments
which essentially does nearest neighbour in the semantic space. TMV-HLP
consistently beats PST in all the results shown in Table~\ref{tab:Comparison-with-stateofart}
thanks to our multi-view embedding. (6) Compared to our TMV-BLP  model \cite{embedding2014ECCV}, the superior results of TMV-HLP shows that the proposed heterogeneous hypergraph is more effective than the homogeneous 2-graphs used in TMV-BLP for zero-shot learning.

Table \ref{tab:Comparison-with-stateofart} also shows that on two
very different datasets: USAA video activity, and CUB fine-grained,
our TMV-HLP significantly outperforms the state-of-the-art alternatives.
In particular, on the more challenging CUB, 47.9\% accuracy is achieved
on 50 classes (chance level 2\%) using the OverFeat feature. Considering
the fine-grained nature and the number of classes, this is even more
impressive than the 80.5\% result on AwA. %It is also noted that again the advantage of our TMV-HLP is even clearer when using the more powerful deep learning feature than the conventional Fisher Vector feature ($\mathcal{F}$). 

\subsubsection{Further evaluations\label{sec:further eva}}

\begin{figure}
\begin{centering}
\includegraphics[scale=0.33]{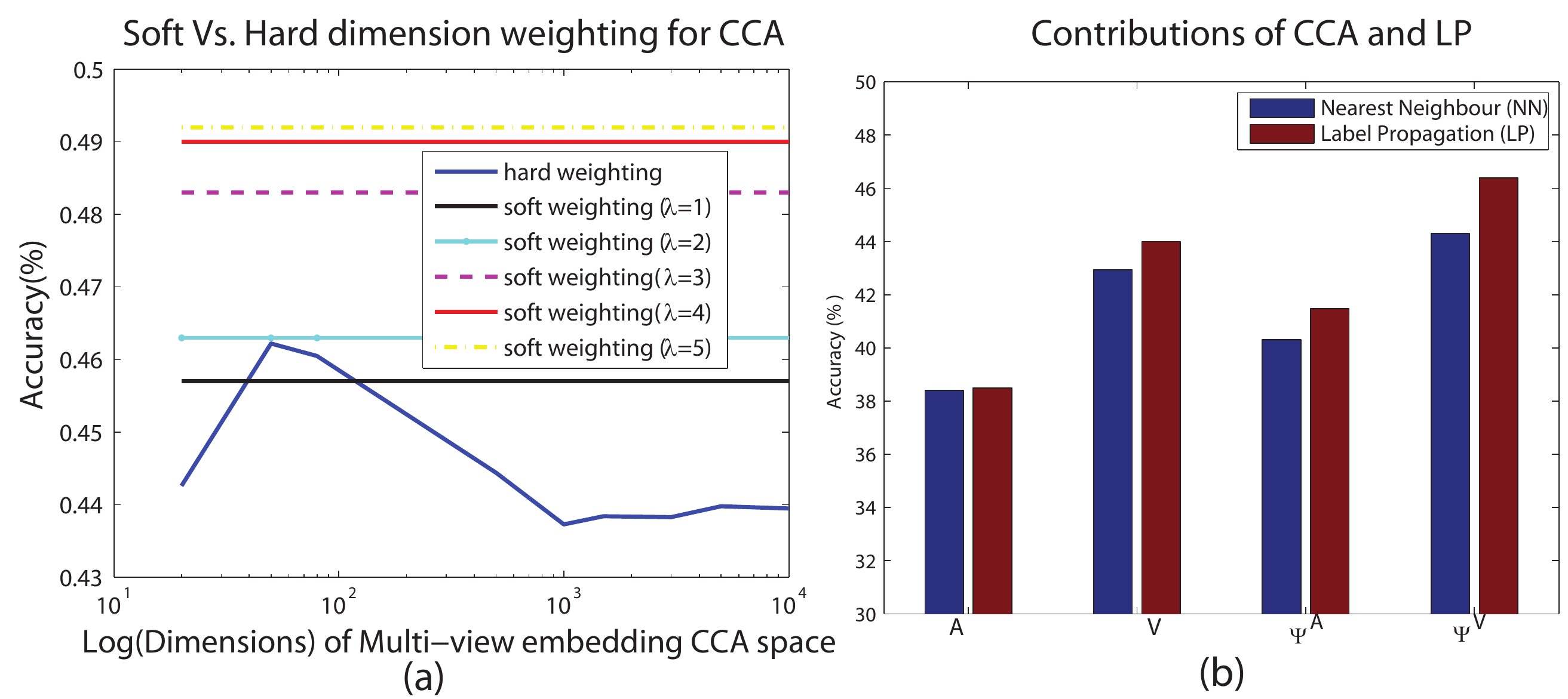}
\par\end{centering}

\protect\caption{\label{fig:CCA-validation} (a) Comparing soft and hard dimension weighting of CCA for AwA. (b) Contributions of CCA and label propagation on  AwA. $\Psi^\mathcal{A}$ and $\Psi^\mathcal{V}$ indicate the subspaces of target data from view $\mathcal{A}$ and $\mathcal{V}$ in $\Gamma$ respectively. Hand-crafted features are used in both experiments. }
\end{figure}

\noindent \textbf{Effectiveness of soft weighting for CCA embedding}\quad{}  
 In this experiment, 
 we compare the soft-weighting (Eq (\ref{eq:ccamapping}))  of  CCA embedding space $\Gamma$ (a strategy adopted in this work) with the conventional hard-weighting   strategy of selecting the number of dimensions for CCA projection.
Fig.~\ref{fig:CCA-validation}(a) shows that  the performance of the hard-weighting CCA depends on the number of projection dimensions selected (blue curve). In contrast,  our soft-weighting strategy uses all dimensions weighted by the CCA eigenvalues, so that the important dimensions are automatically weighted more highly. The result shows that this strategy is clearly better and it is not very sensitive to the weighting parameter $\lambda$, with choices of $\lambda>2$ all working well.

\noindent \textbf{Contributions of individual components}\quad{} 
There are two major components in our ZSL framework: CCA embedding and label propagation. In this experiment we investigate whether both of them contribute to the strong performance. To this end, we compare the ZSL results on AwA with label propagation and without (nearest neighbour) before and after CCA embedding.  In Fig.~\ref{fig:CCA-validation}(b), we can see that: (i) Label propagation always helps regardless whether the views have been embedded using CCA, although its effects are more pronounced after embedding. (ii) Even without label propagation, i.e.~using nearest neighbour for classification, the performance is improved by the CCA embedding. However, the improvement is bigger with label propagation. This result thus suggests that both CCA embedding and label propagation are useful, and our ZSL framework works the best when both are used.

\begin{figure*}[t]
\begin{centering}
\includegraphics[width=0.8\textwidth]{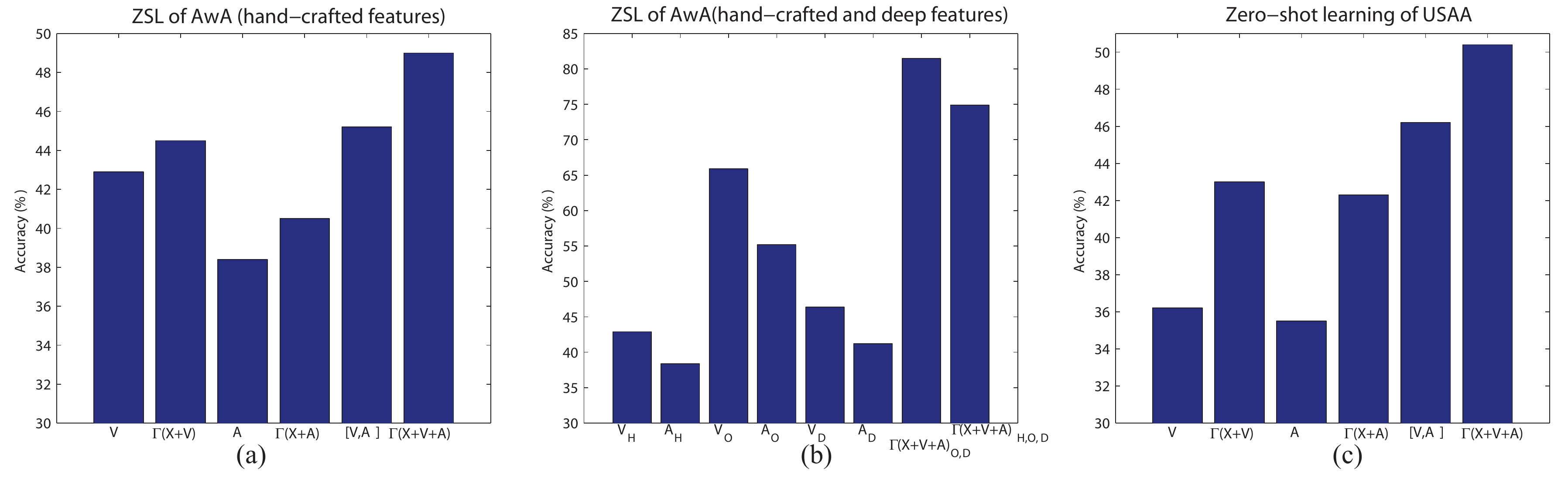} 
\par\end{centering}
\caption{\label{fig:zero-shot-learning-on}Effectiveness of transductive multi-view
embedding. (a) zero-shot learning on AwA using only hand-crafted features;
(b) zero-shot learning on AwA using hand-crafted and deep features
together; (c) zero-shot learning on USAA. $[\mathcal{V},\mathcal{A}]$
indicates the concatenation of semantic word and attribute space vectors.
$\Gamma(\mathcal{X}+\mathcal{V})$ and $\Gamma(\mathcal{X}+\mathcal{A})$
mean using low-level+semantic word spaces and low-level+attribute
spaces respectively to learn the embedding. $\Gamma(\mathcal{X}+\mathcal{V}+\mathcal{A})$
indicates using all $3$ views to learn the embedding. }
\end{figure*}

\noindent \textbf{Transductive multi-view embedding}\quad{}To further 
validate the contribution of our transductive multi-view embedding
space, we split up different views with and without embedding and the
results are shown in Fig.~\ref{fig:zero-shot-learning-on}. In Figs.~\ref{fig:zero-shot-learning-on}(a)
and (c), the hand-crafted feature $\mathcal{H}$ and SIFT, MFCC and
STIP low-level features are used for AwA and USAA respectively, and
we compare $\mathcal{V}$ vs.~$\Gamma(\mathcal{X}+\mathcal{V}$), $\mathcal{A}$ vs.~$\Gamma(\mathcal{X}+\mathcal{A})$ and $[\mathcal{V},\mathcal{A}]$
vs.~$\Gamma(\mathcal{X}+\mathcal{V}+\mathcal{A})$ (see the caption
of Fig.~\ref{fig:zero-shot-learning-on} for definitions). We use
DAP for $\mathcal{A}$ and nearest neighbour for $\mathcal{V}$ and
$[\mathcal{V},\mathcal{A}]$, because the prototypes of $\mathcal{V}$
are not binary vectors so DAP cannot be applied. We use TMV-HLP for
$\Gamma(\mathcal{X}+\mathcal{V})$ and $\Gamma(\mathcal{X}+\mathcal{A})$
respectively. We highlight the following observations: (1) After transductive
embedding, $\Gamma(\mathcal{X}+\mathcal{V}+\mathcal{A})$, $\Gamma(\mathcal{X}+\mathcal{V})$
and $\Gamma(\mathcal{X}+\mathcal{A})$ outperform $[\mathcal{V},\mathcal{A}]$,
$\mathcal{V}$ and $\mathcal{A}$ respectively. This means that the
transductive embedding is helpful whichever semantic space is used
in rectifying the projection domain shift problem by aligning the
semantic views with low-level features. (2) The results of $[\mathcal{V},\mathcal{A}]$
are higher than those of $\mathcal{A}$ and $\mathcal{V}$ individually,
showing that the two semantic views are indeed complementary even
with simple feature level fusion. Similarly, our TMV-HLP on all views
$\Gamma(\mathcal{X}+\mathcal{V}+\mathcal{A})$ improves individual
embeddings $\Gamma(\mathcal{X}+\mathcal{V})$ and $\Gamma(\mathcal{X}+\mathcal{A})$.

\begin{figure*}[ht]
\begin{centering}
\includegraphics[scale=0.4]{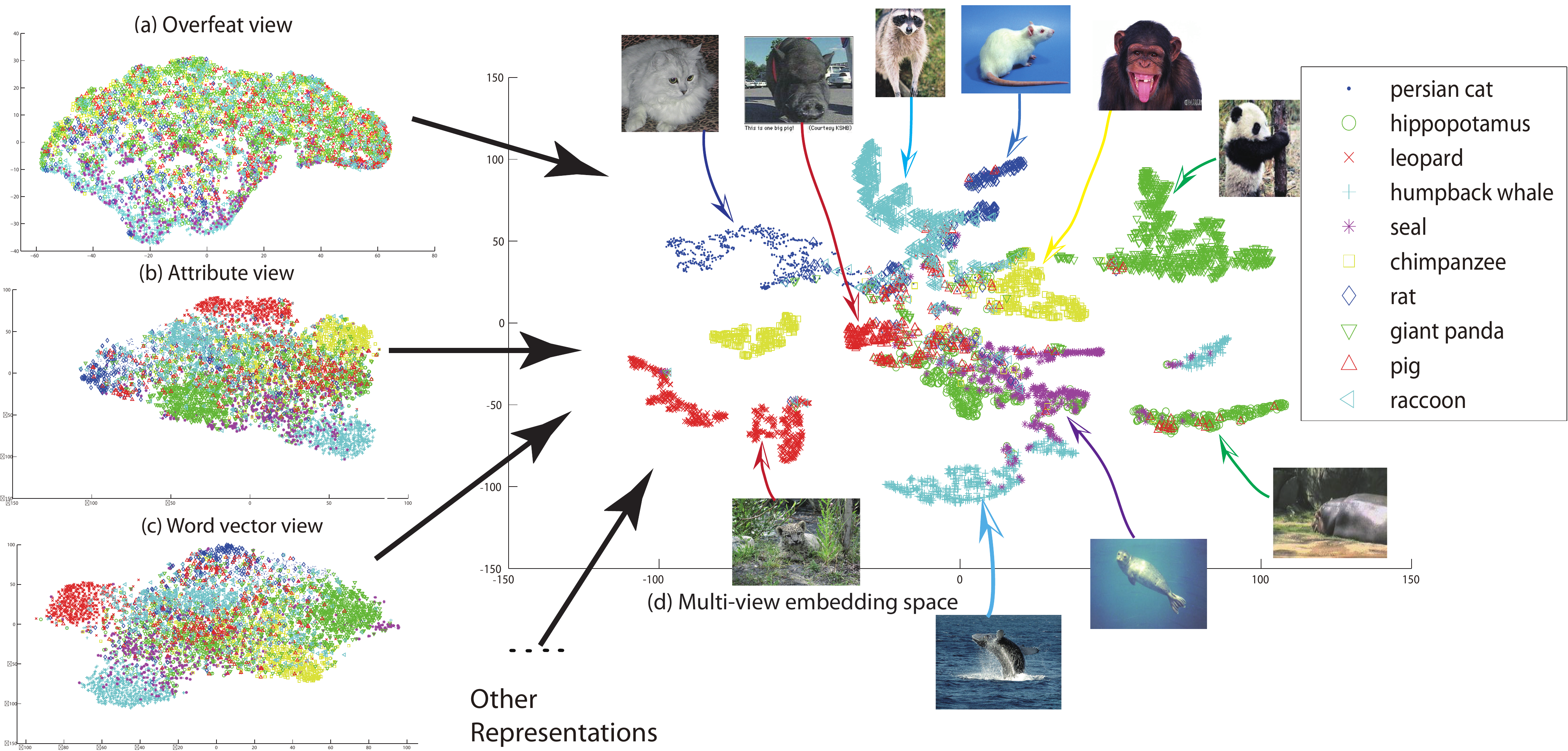} 
\par\end{centering}
\caption{\label{fig:t-SNE-visualisation-of}t-SNE Visualisation of (a) OverFeat
view ($\mathcal{X}_{\mathcal{O}}$), (b) attribute view ($\mathcal{A}_{\mathcal{O}}$),
(c) word vector view ($\mathcal{V}_{\mathcal{O}}$), and (d) transition
probability of pairwise nodes computed by Eq (\ref{eq:transition_probability})
of TMV-HLP in ($\Gamma(\mathcal{X}+\mathcal{A}+\mathcal{V})_{\mathcal{O},\mathcal{D}}$).
The unlabelled target classes are much more separable in (d).}
\end{figure*}

\noindent \textbf{Embedding deep learning feature views also helps}\quad{}In Fig.~\ref{fig:zero-shot-learning-on}(b)
three different low-level features are considered for AwA: hand-crafted
($\mathcal{H}$), OverFeat ($\mathcal{O}$) and DeCAF features ($\mathcal{D}$).
The zero-shot learning results of each individual space are indicated
as $\mathcal{V}_{\mathcal{H}}$, $\mathcal{A}_{\mathcal{H}}$, $\mathcal{V}_{\mathcal{O}}$,
$\mathcal{A}_{\mathcal{O}}$, $\mathcal{V}_{\mathcal{D}}$, $\mathcal{A}_{\mathcal{D}}$
in Fig.~\ref{fig:zero-shot-learning-on}(b) and we observe that $\mathcal{V}_{\mathcal{O}}>\mathcal{V}_{\mathcal{D}}>\mathcal{V}_{\mathcal{H}}$
and $\mathcal{A}_{\mathcal{O}}>\mathcal{A}_{\mathcal{D}}>\mathcal{A}_{\mathcal{H}}$.
That is OverFeat $>$ DeCAF $>$ hand-crafted features. It is widely
reported that deep features have better performance than `hand-crafted'
features on many computer vision benchmark datasets \cite{2014arXiv1405.3531C,sermanet-iclr-14}.
What is interesting to see here is that OverFeat $>$ DeCAF since
both are based on the same Convolutional Neural Network (CNN) model
of \cite{KrizhevskySH12}. Apart from implementation details, one
significant difference is that DeCAF is pre-trained by ILSVRC2012
while OverFeat by ILSVRC2013 which contains more animal classes meaning
better (more relevant) features can be learned. It is also worth pointing
out that: (1) With both OverFeat and DeCAF features, the number of
views to learn an embedding space increases from $3$ to $9$; and our
results suggest that the more views, the better chance to solve the
domain shift problem and the data become more separable as different
views contain complementary information. % Strictly speaking the deep learning features OverFeat and DeCAF are not low-level features -- with the deep architecture learned in a supervised manner, it is more %appropriate to consider those features as semantic features at the object level. This explains why embedding both $\mathcal{O}$,
%$\mathcal{D}$ and their semantic projections further boosting the results -- more semantic views have a better chance to solve the domain shift problem and the data become %more separable as different views contain complementary information.
(2) Figure~\ref{fig:zero-shot-learning-on}(b) shows that when all
9 available views ($\mathcal{X}_{\mathcal{H}}$, $\mathcal{V}_{\mathcal{H}}$,
$\mathcal{A}_{\mathcal{H}}$, $\mathcal{X}_{\mathcal{D}}$, $\mathcal{V}_{\mathcal{D}}$,
$\mathcal{A}_{\mathcal{D}}$, $\mathcal{X}_{\mathcal{O}}$, $\mathcal{V}_{\mathcal{O}}$
and $\mathcal{A}_{\mathcal{O}}$) are used for embedding, the result
is significantly better than those from each individual view. Nevertheless,
it is lower than that obtained by embedding views ($\mathcal{X}_{\mathcal{D}}$,
$\mathcal{V}_{\mathcal{D}}$, $\mathcal{A}_{\mathcal{D}}$, $\mathcal{X}_{\mathcal{O}}$,
$\mathcal{V}_{\mathcal{O}}$ and $\mathcal{A}_{\mathcal{O}}$). This
suggests that view selection may be required when a large number of
views are available for learning the embedding space.

%of three views ($\mathcal{X}_{\mathcal{H}}$,
%$\mathcal{V}_{\mathcal{H}}$ and $\mathcal{A}_{\mathcal{H}}$) generated
%by hand-crafted features are significantly lower than the other views,  the
%results of only using deep feature views ($\mathcal{X}_{\mathcal{D}}$,
%$\mathcal{V}_{\mathcal{D}}$, $\mathcal{A}_{\mathcal{D}}$, $\mathcal{X}_{\mathcal{O}}$,
%$\mathcal{V}_{\mathcal{O}}$ and $\mathcal{A}_{\mathcal{O}}$) achieves the best performance (denoted by $\Gamma(\mathcal{X}+\mathcal{A}+\mathcal{V})_{\mathcal{O},%\mathcal{D}}$).

\noindent \textbf{Embedding makes target classes more separable}\quad{}We
employ t-SNE \cite{tsne} to visualise the space $\mathcal{X}_{\mathcal{O}}$,
$\mathcal{V}_{\mathcal{O}}$, $\mathcal{A}_{\mathcal{O}}$ and $\Gamma(\mathcal{X}+\mathcal{A}+\mathcal{V})_{\mathcal{O},\mathcal{D}}$
in Fig.~\ref{fig:t-SNE-visualisation-of}. It shows that even in
the powerful OverFeat view, the 10 target classes are heavily overlapped
(Fig.~\ref{fig:t-SNE-visualisation-of}(a)). It gets better in the
semantic views (Figs.~\ref{fig:t-SNE-visualisation-of}(b) and (c)).
However, when all 6 views are embedded, all classes are clearly separable
(Fig.~\ref{fig:t-SNE-visualisation-of}(d)).

\noindent \textbf{Running time}\quad{}In practice, for the AwA dataset
with hand-crafted features, our pipeline takes less than $30$ minutes
to complete the zero-shot classification task (over $6,180$ images)
using a six core $2.66$GHz CPU platform. This includes the time for
multi-view CCA embedding and label propagation using our heterogeneous
hypergraphs.

%More evaluation of our method for zero-shot and N-shot learning can be found in the appendix.

\subsection{Annotation and beyond}

In this section we evaluate our multi-view embedding space for the
conventional and novel annotation tasks introduced in Sec.~\ref{sec:Annotation-and-Beyond}.

\noindent \textbf{Instance annotation by attributes}\quad{}To quantify
the annotation performance, we predict attributes/annotations for
each target class instance for USAA, which has the largest instance
level attribute variations among the three datasets. We employ two
standard measures: mean average precision (mAP) and F-measure (FM)
between the estimated and true annotation list. Using our multi-view
embedding space, our method (FM: $0.341$, mAP: $0.355$) outperforms
significantly the baseline of directly estimating $\mathbf{y}_{u}^{\mathcal{A}}=f^{\mathcal{A}}(\mathbf{x}_{u})$
(FM: $0.299$, mAP: $0.267$).

\begin{table}[ht]
\begin{centering}
\begin{tabular}{|c|c|c|}
\hline 
AwA &  & Attributes\tabularnewline
\hline 
\hline 
\multirow{2}{*}{pc} & T-5 & active, \textbf{furry, tail, paws, ground.}\tabularnewline
\cline{2-3} 
 & B-5 & swims, hooves, long neck, horns, arctic\tabularnewline
\hline 
\multirow{2}{*}{hp} & T-5 & \textbf{old world, strong, quadrupedal}, fast, \textbf{walks}\tabularnewline
\cline{2-3} 
 & B-5 & red, plankton, skimmers, stripes, tunnels\tabularnewline
\hline 
\multirow{2}{*}{lp} & T-5 & \textbf{old world, active, fast, quadrupedal, muscle}\tabularnewline
\cline{2-3} 
 & B-5 & plankton, arctic, insects, hops, tunnels\tabularnewline
\hline 
\multirow{2}{*}{hw} & T-5 & \textbf{fish}, \textbf{smart, fast, group, flippers}\tabularnewline
\cline{2-3} 
 & B-5 & hops, grazer, tunnels, fields, plains\tabularnewline
\hline 
\multirow{2}{*}{seal} & T-5 & \textbf{old world, smart}, \textbf{fast}, chew teeth, \textbf{strong}\tabularnewline
\cline{2-3} 
 & B-5 & fly, insects, tree, hops, tunnels\tabularnewline
\hline 
\multirow{2}{*}{cp} & T-5 & \textbf{fast, smart, chew teeth, active}, brown\tabularnewline
\cline{2-3} 
 & B-5 & tunnels, hops, skimmers, fields, long neck\tabularnewline
\hline 
\multirow{2}{*}{rat} & T-5 & \textbf{active, fast, furry, new world, paws}\tabularnewline
\cline{2-3} 
 & B-5 & arctic, plankton, hooves, horns, long neck\tabularnewline
\hline 
\multirow{2}{*}{gp} & T-5 & \textbf{quadrupedal}, active, \textbf{old world}, \textbf{walks},
\textbf{furry}\tabularnewline
\cline{2-3} 
 & B-5 & tunnels, skimmers, long neck, blue, hops\tabularnewline
\hline 
\multirow{2}{*}{pig} & T-5 & \textbf{quadrupedal}, \textbf{old world}, \textbf{ground}, furry,
\textbf{chew teeth}\tabularnewline
\cline{2-3} 
 & B-5 & desert, long neck, orange, blue, skimmers\tabularnewline
\hline 
\multirow{2}{*}{rc} & T-5 & \textbf{fast}, \textbf{active}, \textbf{furry}, \textbf{quadrupedal},
\textbf{forest}\tabularnewline
\cline{2-3} 
 & B-5 & long neck, desert, tusks, skimmers, blue\tabularnewline
\hline 
\end{tabular}
\par\end{centering}

\caption{Zero-shot description of 10 AwA target classes. $\Gamma$ is learned
using 6 views ($\mathcal{X}_{\mathcal{D}}$, $\mathcal{V}_{\mathcal{D}}$,
$\mathcal{A}_{\mathcal{D}}$, $\mathcal{X}_{\mathcal{O}}$, $\mathcal{V}_{\mathcal{O}}$
and $\mathcal{A}_{\mathcal{O}}$). The true positives are highlighted
in bold. pc, hp, lp, hw, cp, gp, and rc are short for Persian cat,
hippopotamus, leopard, humpback whale, chimpanzee, giant panda, and
raccoon respectively. T-5/B-5 are the top/bottom 5 attributes predicted
for each target class.}
\label{fig:ZeroShotDescription} 
\end{table}

\begin{table*}[ht]
\begin{centering}
\begin{tabular}{|c|c|c|}
\hline 
(a) Query by GT attributes of  & Query via embedding space  & Query attribute words in word space\tabularnewline
\hline 
\hline 
graduation party  & \textbf{party}, \textbf{graduation}, audience, caucus  & cheering, proudly, dressed, wearing\tabularnewline
\hline 
music\_performance  & \textbf{music}, \textbf{performance}, musical, heavy metal  & sing, singer, sang, dancing\tabularnewline
\hline 
wedding\_ceremony  & \textbf{wedding\_ceremony}, wedding, glosses, stag  & nun, christening, bridegroom, \textbf{wedding\_ceremony}\tabularnewline
\hline 
\end{tabular}
\par\end{centering}

\begin{centering}
\begin{tabular}{|c|c|}
\hline 
(b) Attribute query  & Top ranked words\tabularnewline
\hline 
\hline 
wrapped presents  & music; performance; solo\_performances; performing\tabularnewline
\hline 
+small balloon  & wedding; wedding\_reception; birthday\_celebration; birthday\tabularnewline
\hline 
+birthday song +birthday caps  & \textbf{birthday\_party}; prom; wedding reception\tabularnewline
\hline 
\end{tabular}
\par\end{centering}
\caption{\label{fig:ZAL_Task}Zero prototype learning on USAA. (a) Querying
classes by groundtruth (GT) attribute definitions of the specified
classes. (b) An incrementally constructed attribute query for the
birthday\_party class. Bold indicates true positive.}
\end{table*}

\noindent \textbf{Zero-shot description}\textit{\quad{}}\textit{\emph{In
this}} task, we explicitly infer the attributes corresponding to a
specified novel class, given only the textual name of that class without
seeing any visual samples. Table~\ref{fig:ZeroShotDescription} illustrates
this for AwA. Clearly most of the top/bottom $5$ attributes predicted
for each of the 10 target classes are meaningful (in the ideal case,
all top $5$ should be true positives and all bottom $5$ true negatives).
Predicting the top-$5$ attributes for each class gives an F-measure of $0.236$. 
In comparison, if we directly
select the $5$ nearest attribute name projection to the class name
projection (prototype) in the word space, the F-measure is
$0.063$, demonstrating the importance of learning the multi-view
embedding space. In addition to providing a method to automatically
-- rather than manually -- generate an attribute ontology, this task
is interesting because even a human could find it very challenging
(effectively a human has to list the attributes of a class which he
has never seen or been explicitly taught about, but has only seen
mentioned in text).

\noindent \textbf{Zero prototype learning}\quad{}In this task we
attempt the reverse of the previous experiment: inferring a class
name given a list of attributes. Table \ref{fig:ZAL_Task} illustrates
this for USAA. Table \ref{fig:ZAL_Task}(a) shows queries by the groundtruth
attribute definitions of some USAA classes and the top-4 ranked
list of classes returned. The estimated class names of each attribute
vector are reasonable -- the top-4 words are either the class name
or related to the class name. A baseline is to use the textual names
of the attributes projected in the word space (summing their
word vectors) to search for the nearest classes in word space, instead
of the embedding space. Table \ref{fig:ZAL_Task}(a) shows that the
predicted classes in this case are reasonable, but significantly
worse than querying via the embedding space. To quantify this we evaluate
the average rank of the true name for each USAA class when queried
by its attributes. For querying by embedding space, the average rank
 is an impressive $2.13$ (out of $4.33$M words
with a chance-level rank of $2.17$M), compared with the average rank
of $110.24$ by directly querying word space \cite{wordvectorICLR}
with textual descriptions of the attributes. Table \ref{fig:ZAL_Task}(b)
shows an example of ``incremental'' query using the ontology definition
of birthday party \cite{yanweiPAMIlatentattrib}. We first query the
`wrapped presents' attribute only, followed by adding `small
balloon' and all other attributes (`birthday songs and `birthday
caps'). The changing list of top ranked retrieved words intuitively
reflects the expectation of the combinatorial meaning of the attributes.

\section{Conclusions}

We identified the challenge of projection domain shift in zero-shot
learning and presented a new framework to solve it by rectifying the
biased projections in a multi-view embedding space. We also proposed
a novel label-propagation algorithm TMV-HLP based on heterogeneous
across-view hypergraphs. TMV-HLP synergistically exploits multiple
intermediate semantic representations, as well as the manifold structure
of unlabelled target data to improve recognition in a unified way
for zero shot, N-shot and zero+N shot learning tasks. As a result
we achieved state-of-the-art performance on the challenging AwA, CUB
and USAA datasets. Finally, we demonstrated that our framework enables
novel tasks of relating textual class names and their semantic attributes.

A number of directions have been identified for future work. First,
we employ CCA for learning the embedding space. Although
it works well, other embedding frameworks can be considered (e.g.~\cite{DBLP:conf/iccv/WangHWWT13}).
In the current pipeline, low-level features are first
projected onto different semantic views before embedding.
It should be possible to develop a unified embedding framework to combine these
two steps. Second, under a realistic lifelong learning
setting \cite{chen_iccv13}, an unlabelled data point could either
belong to a seen/auxiliary category or
an unseen class. An ideal framework should be able to classify both seen and unseen classes \cite{RichardNIPS13}.
Finally, our results suggest that more views, either manually defined
(attributes), extracted from a linguistic corpus (word space),
or learned from visual data (deep features), can potentially give
rise to better embedding space. More investigation is needed on how to systematically design and select semantic views for
embedding.

\bibliographystyle{abbrv}
\bibliography{ref-phd1}

\begin{IEEEbiography}[{\includegraphics[width=1in,height=1.25in,clip,keepaspectratio]{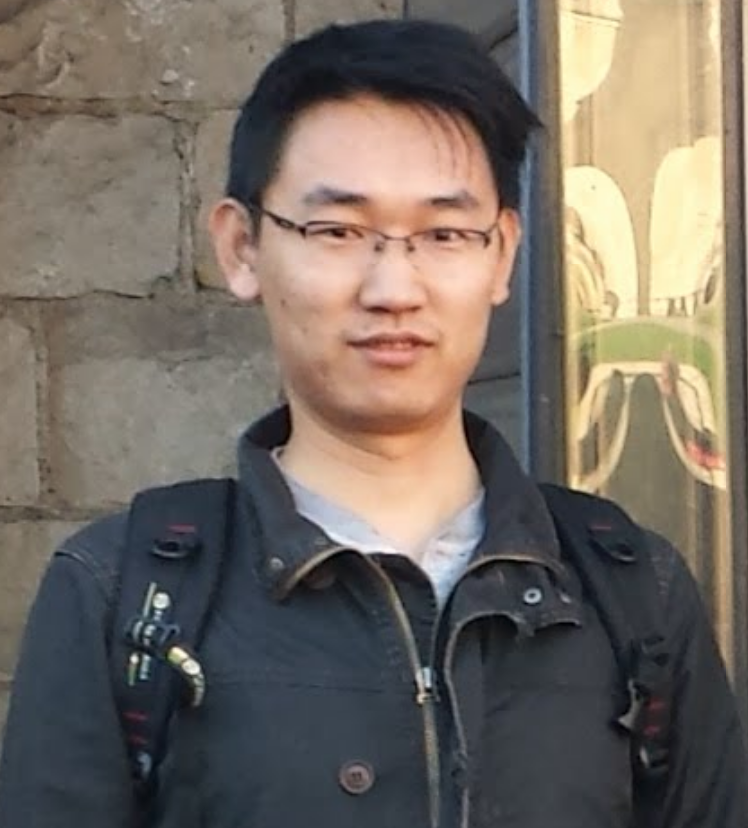}}]{Yanwei Fu} received the PhD degree from Queen Mary University of London in 2014, and the MEng degree in the Department of Computer Science \& Technology at Nanjing University in 2011, China. He is a Post-doctoral researcher with Leonid Sigal in Disney Research, Pittsburgh, which is co-located with Carnegie Mellon University. His research interests include attribute learning for image and video understanding, robust learning to rank and large-scale video summarization.
\end{IEEEbiography}

\begin{IEEEbiography}
[{\includegraphics[width=1in,height=1.25in,clip,keepaspectratio]{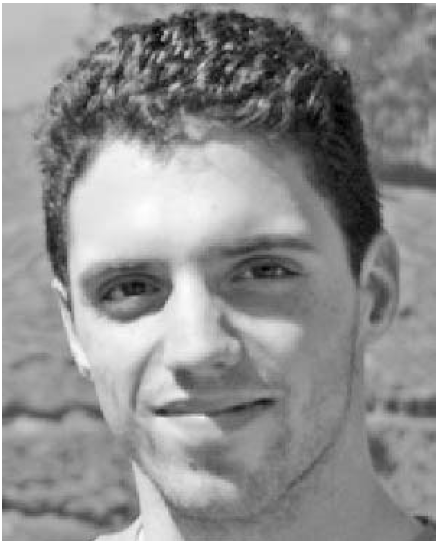}}]{Timothy M. Hospedales} received  the  PhD degree in neuroinformatics from the University of Edinburgh in 2008. He is currently a lecturer (assistant professor) of computer science at Queen Mary University of London. His research interests include probabilistic modelling and machine learning applied variously to problems in computer vision, data mining, interactive learning, and neuroscience. He has published more than 20  papers  in  major  international journals and conferences. He is a member of the IEEE.
\end{IEEEbiography}

\begin{IEEEbiography}[{\includegraphics[width=1in,height=1.25in,clip,keepaspectratio]{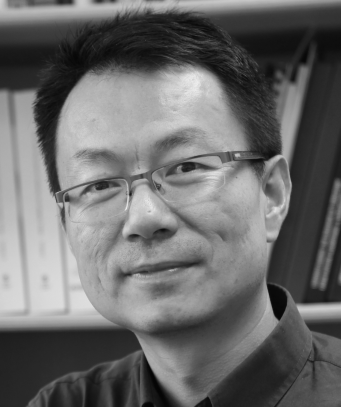}}]{Tao Xiang} received the PhD degree in electrical and computer engineering from the National University of Singapore in 2002. He is currently a reader (associate professor) in the School of Electronic Engineering and Computer Science, Queen Mary University of London. His research interests include computer vision and  machine learning. He has published over 100 papers in international journals and conferences and co-authored a book, Visual Analysis of Behaviour: From Pixels to Semantics.
\end{IEEEbiography}

\begin{IEEEbiography}[{\includegraphics[width=1in,height=1.25in,clip,keepaspectratio]{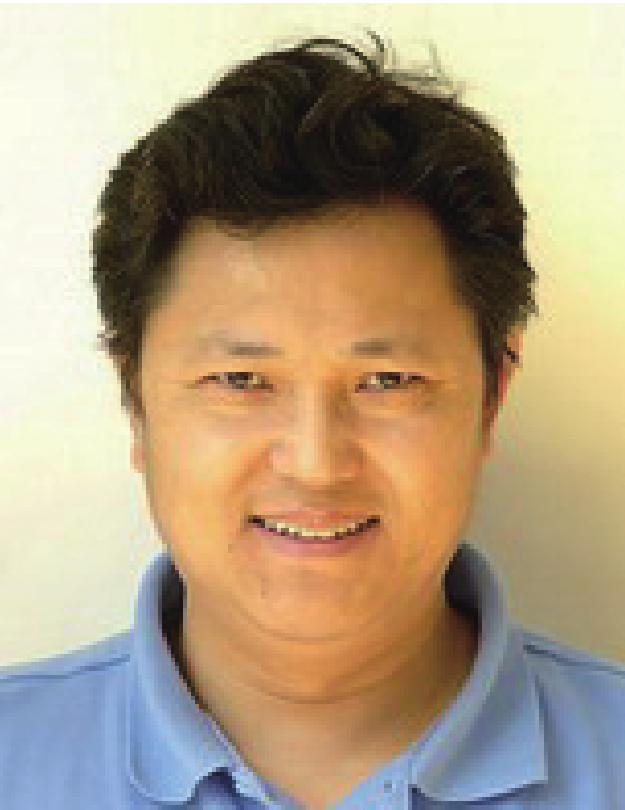}}]{Shaogang Gong}  is Professor of Visual Computation at Queen Mary University of London, a Fellow of the Institution of Electrical Engineers and a Fellow of the British Computer Society. He received his D.Phil in computer vision from Keble College, Oxford University in 1989. His research interests include computer vision, machine learning and video analysis.
\end{IEEEbiography}

\section*{Supplementary Material}

\section{Further Evaluations on Zero-Shot Learning}

\begin{figure}[h]
\centering{}\includegraphics[width=0.5\textwidth]{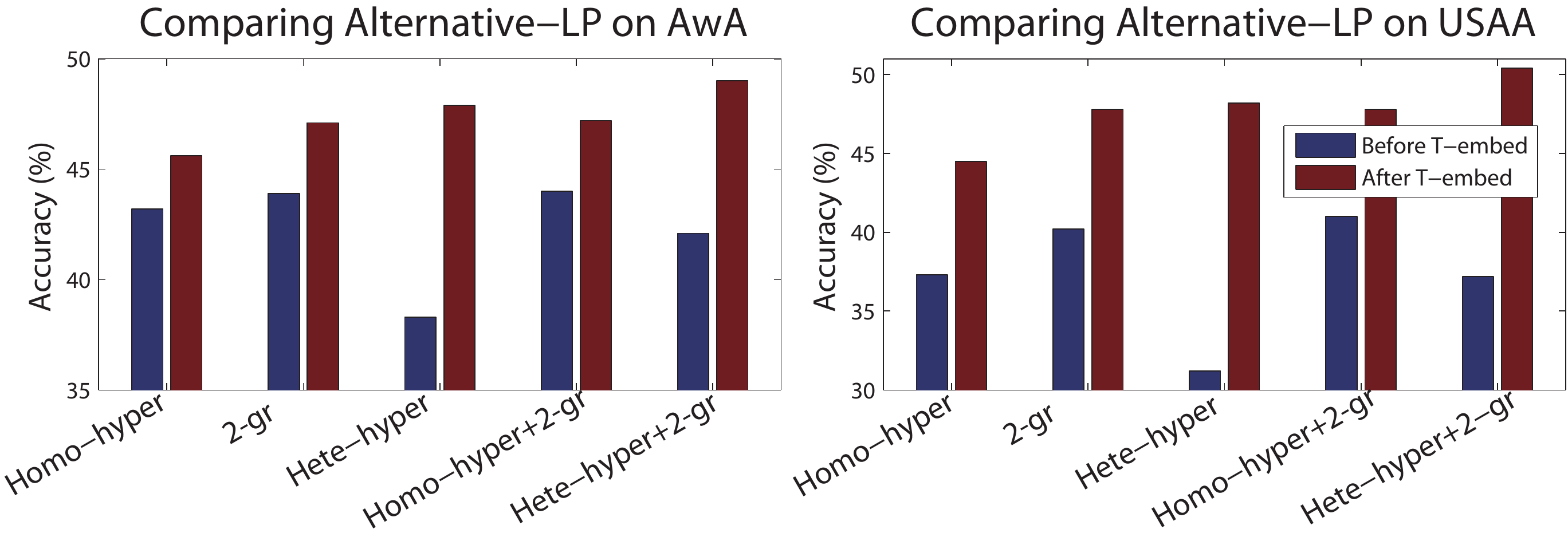}
\caption{\label{fig:Comparing-alternative-LP}Comparing alternative label propagation
methods using different graphs before and after transductive embedding (T-embed).  The methods
are detailed in text.}
\end{figure}

\subsection{\noindent Heterogeneous hypergraph vs. other graphs\quad{}}

\noindent Apart from transductive multi-view embedding, another major
contribution of this paper is a novel label propagation method based
on heterogeneous hypergraph. To evaluate the effectiveness of our
hypergraph label propagation, we compare with a number of alternative
label propagation methods using other graph models. More specifically,
within each view, two alternative graphs can be constructed: 2-graphs
which are used in the classification on multiple graphs (C-MG) model
\cite{Zhou2007ICML}, and conventional homogeneous hypergraph formed
in each single view \cite{Zhou06learningwith,fu2010summarize,DBLP:journals/corr/LiLSDH13}.
Since hypergraphs are typically combined with 2-graphs, we have 5
different multi-view graph models: \emph{2-gr} (2-graph in each view),
\emph{Homo-hyper} (homogeneous hypergraph in each view), \emph{Hete-hyper}
(our heterogeneous hypergraph across views), \emph{Homo-hyper+2-gr}
(homogeneous hypergraph combined with 2-graph in each view), and \emph{Hete-hyper+2-gr}
(our heterogeneous hypergraph combined with 2-graph, as in our TMV-HLP).
In our experiments, the same random walk label propagation algorithm
is run on each graph in AwA and USAA before and after transductive
embedding to compare these models.

From the results in Fig.~\ref{fig:Comparing-alternative-LP}, we
observe that: (1) The graph model used in our TMV-HLP (\emph{Hete-hyper+2-gr})
yields the best performance on both datasets. (2) All graph models
benefit from the embedding. In particular, the performance of our
heterogeneous hypergraph degrades drastically without embedding. This
is expected because before embedding, nodes in different views are
not aligned; so forming meaningful hyperedges across views is not
possible. (3) Fusing hypergraphs with 2-graphs helps -- one has the
robustness and the other has the discriminative power, so it makes
sense to combine the strengths of both. (4) After embedding, on its
own, heterogeneous graphs are the best while homogeneous hypergraphs
(\emph{Homo-hyper}) are worse than 2-gr indicating that the discriminative
power by 2-graphs over-weighs the robustness of homogeneous hypergraphs.

\begin{figure}[tph]
\centering{}\includegraphics[scale=0.45]{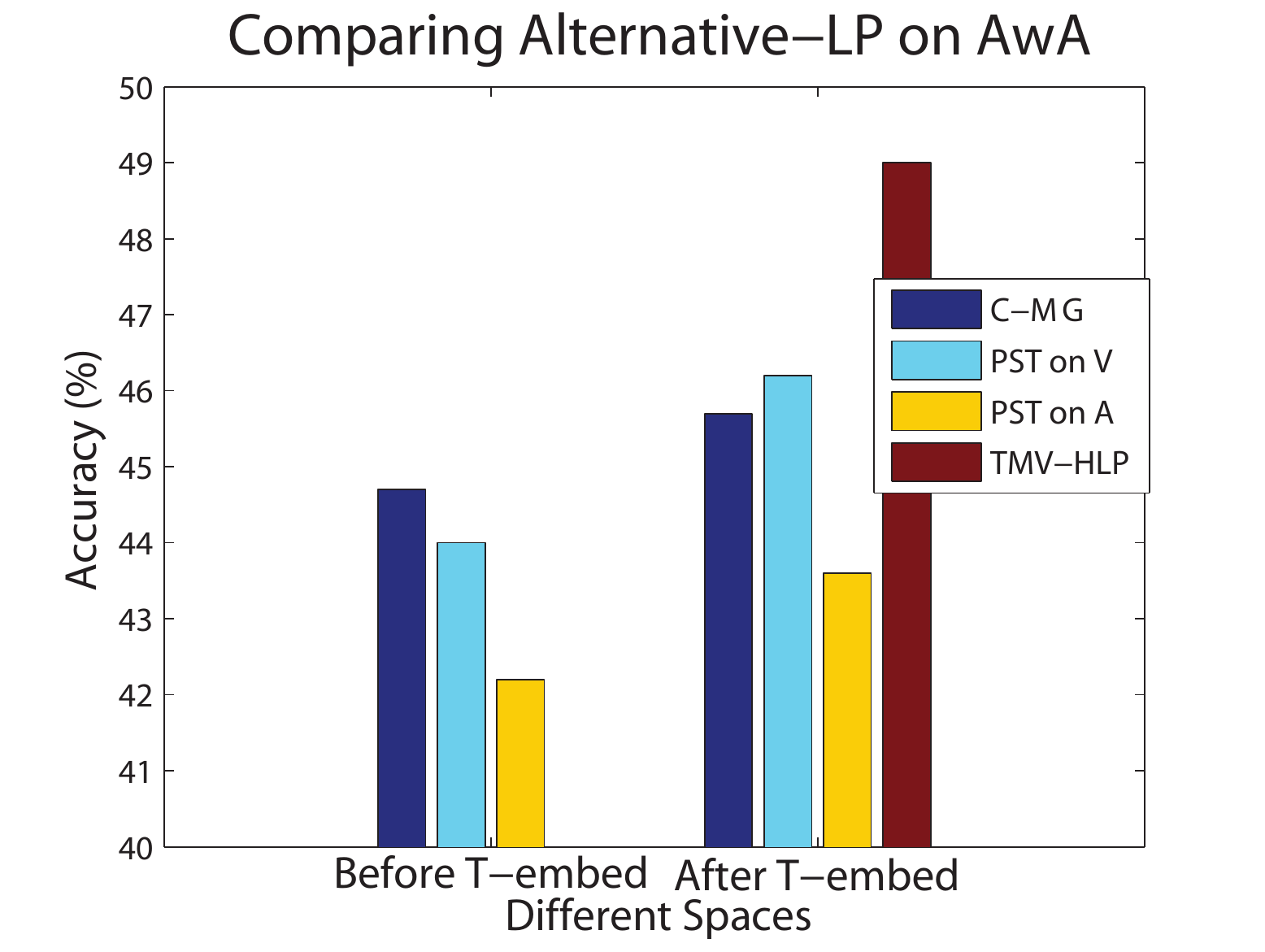}\caption{\label{fig:Comparing-C-MG-and}Comparing our method with the alternative C-MG and PST methods before and after
transductive embedding}
\end{figure}

\subsection{Comparing with other transductive methods}

 Rohrbarch et al.~\cite{transferlearningNIPS}  employed an alternative transductive method, termed PST, for zero-shot learning. We compare
with PST for  zero-shot learning here and N-shot learning
 in Sec.~\ref{sec:n-shot}.

Specifically, we
use AwA dataset with hand-crafted features, semantic word vector $\mathcal{V}$,
and semantic attribute $\mathcal{A}$. We compare with PST
\cite{transferlearningNIPS} as well as the graph-based
semi-supervised learning methods C-MG \cite{Zhou2007ICML} (not originally designed but can be used for zero-shot learning) before and after our transductive embedding
(T-embed). We use equal weights for each graph for C-MG and the same
parameters from \cite{transferlearningNIPS} for PST. From Fig.~\ref{fig:Comparing-C-MG-and},
we make the following conclusions: (1) TMV-HLP in our embedding space
outperforms both alternatives. (2) The embedding also improves both
C-MG and PST, due to alleviated projection domain shift via aligning
the semantic projections and low-level features.

The reasons for TMV-HLP outperforming PST include: (1) Using multiple semantic
views (PST is defined only on one), (2) Using hypergraph-based label
propagation (PST uses only conventional 2-graphs), (3) Less dependence
on good quality initial labelling heuristics required by PST -- our
TMV-HLP uses the trivial initial labels (each prototype labelled according
to its class as in Eq (17) in the main manuscript).

\subsection{How many target samples are needed for learning multi-view embedding?}

\begin{figure}[tbph]
\centering{}\includegraphics[scale=0.5]{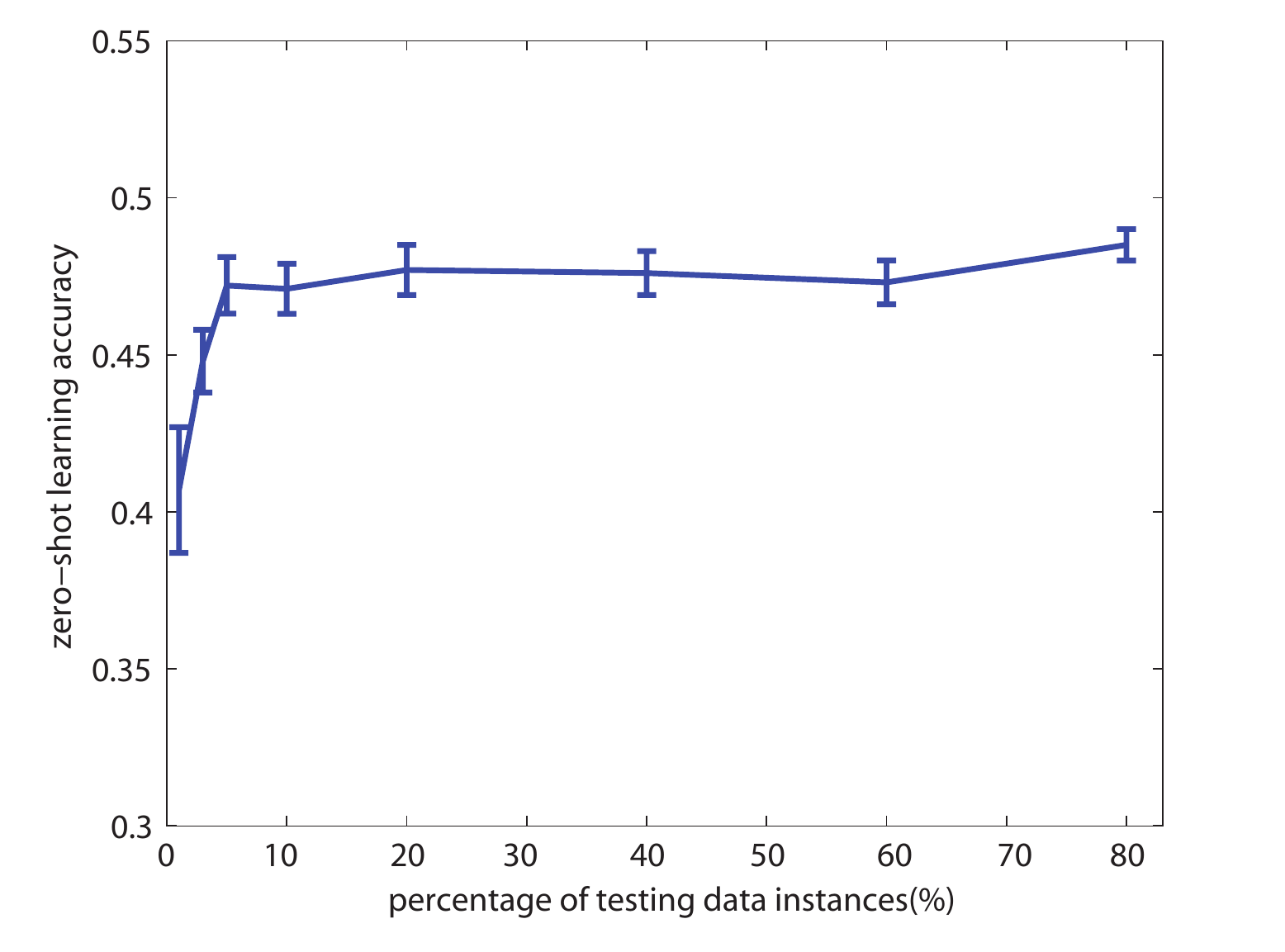}\caption{Influence of varying the number of unlabelled target class samples used to learn the CCA space.}
\label{fig:samples for cca}
\end{figure}

In our paper, we use all the target class samples to construct the transductive
embedding CCA space. Here we investigate how many samples are required to construct a reasonable
embedding.  We use hand-crafted features (dimension:
$10,925$) of the AwA dataset with semantic word vector $\mathcal{V}$
(dimension: $1,000$) and semantic attribute $\mathcal{A}$ (dimension:
$85$) to construct the CCA space. We randomly select $1\%$, $3\%$,
$5\%$, $20\%$, $40\%$, $60\%$, and $80\%$ of the unlabelled target class
instances to  construct the CCA space for zero-shot learning using our TMV-HLP.
Random sampling is repeated $10$ times. The results shown in Fig.~\ref{fig:samples for cca} below demonstrate that only 5\% of the full set of samples
(300 in the case of AwA) are sufficient to learn a good embedding
space. 

\subsection{Can the embedding space be learned using the auxiliary dataset? }

\begin{figure}[tbh]
\begin{centering}
\includegraphics[scale=0.4]{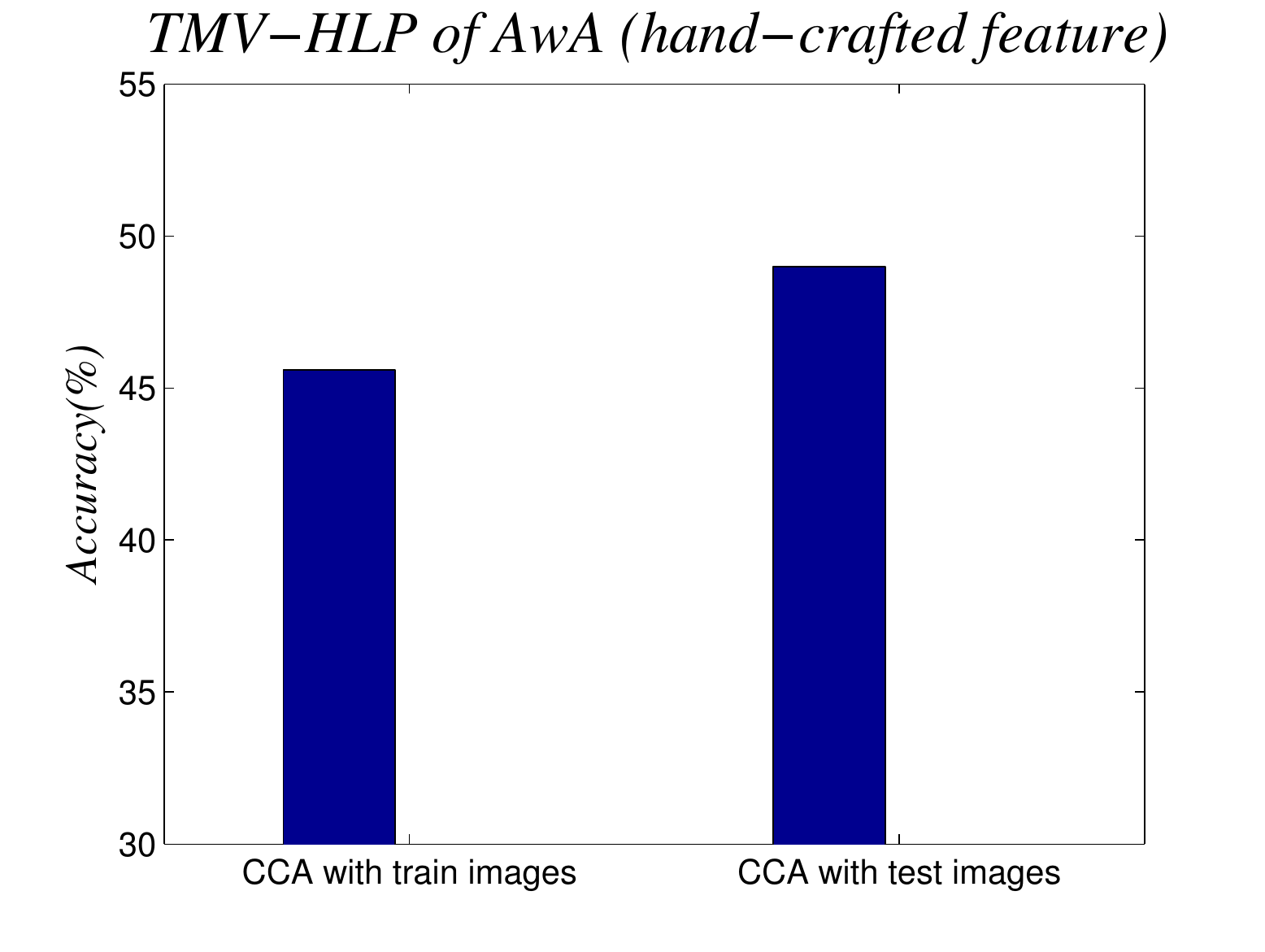}
\par\end{centering}

\caption{\label{fig:CCA-space-trained}Comparing ZSL performance using CCA
learned from auxiliary data versus unlabelled target data. }
\end{figure}

Since we aim to rectify the projection domain shift problem for the target data, the multi-view embedding space is learned transductively using the target dataset. One may ask whether the multi-view embedding space learned using the auxiliary dataset can be of any use for the zero-shot classification of the target class samples. To answer this question,
we conduct experiments by using the hand-crafted features (dimension:
$10,925$) of AwA dataset with semantic word vector $\mathcal{V}$
(dimension: $1,000$) and semantic attribute $\mathcal{A}$ (dimension:
$85$). Auxiliary data from AwA are now used to learn the multi-view
CCA, and we then project the testing data into this CCA space.  
We compare the results of our TMV-HLP on AwA using the CCA spaces learned
from the auxiliary dataset versus unlabelled target dataset in Fig.~\ref{fig:CCA-space-trained}. It can be seen that the CCA embedding space learned using the  auxiliary dataset gives
reasonable performance; however, it does not perform as well as our
method which learned the multi-view embedding space transductively using target class samples. This
is likely due to not observing, and thus not being able to learn to rectify,
the projection domain shift.

\subsection{Qualitative results}

\begin{figure}[h!]
\begin{centering}
\includegraphics[scale=0.5]{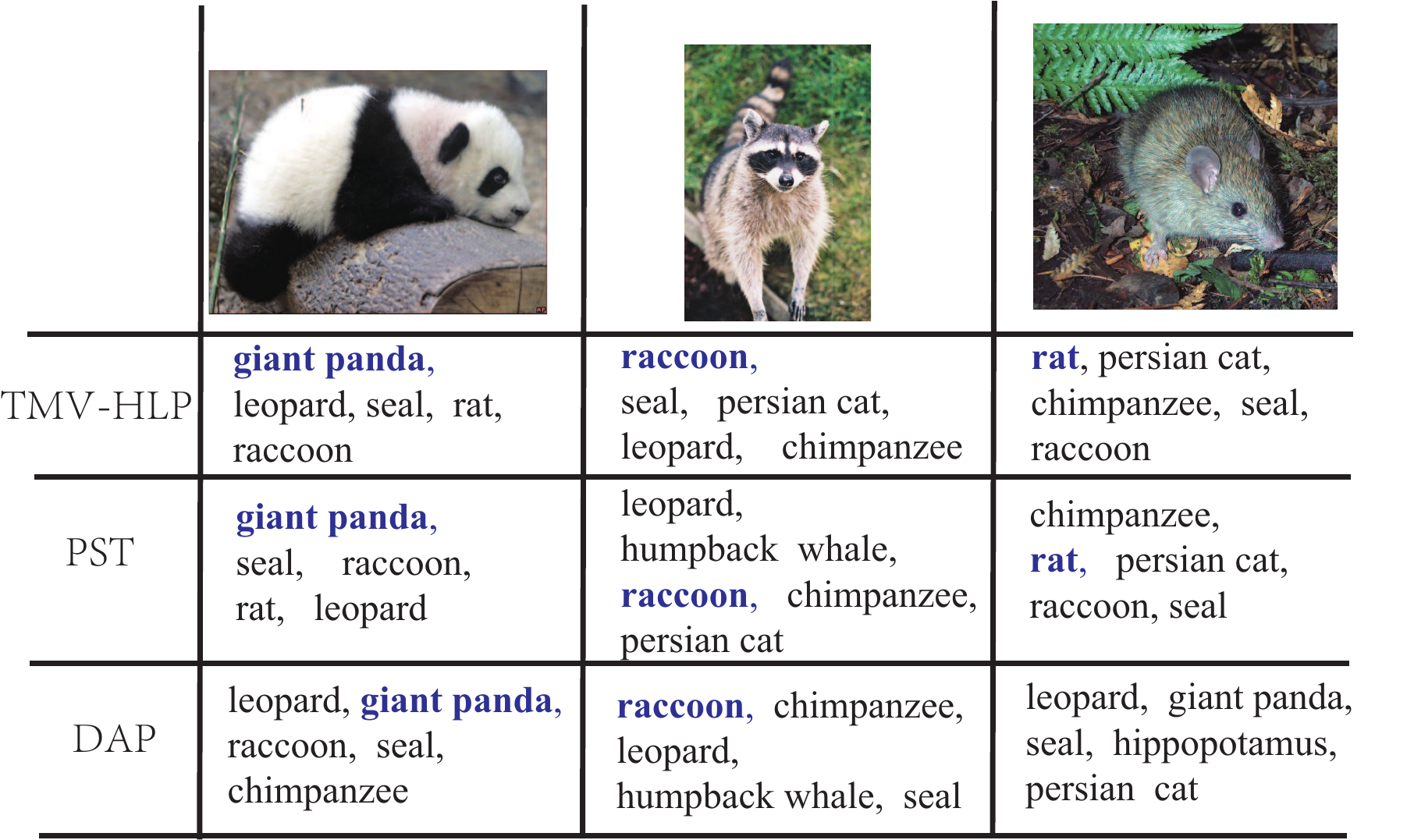}
\par\end{centering}
\caption{\label{fig:QualitativeZSLAwA}Qualitative results for zero-shot learning
on AwA. Bold indicates correct class names.}
\end{figure}

Figure \ref{fig:QualitativeZSLAwA} shows some qualitative results for
zero-shot learning on AwA in terms of top 5 most likely classes predicted
for each image. It shows that our TMV-HLP produces more reasonable ranked list of classes
for each image, comparing to DAP and PST.

\begin{figure*}[ht!]
\begin{centering}
\includegraphics[scale=0.4]{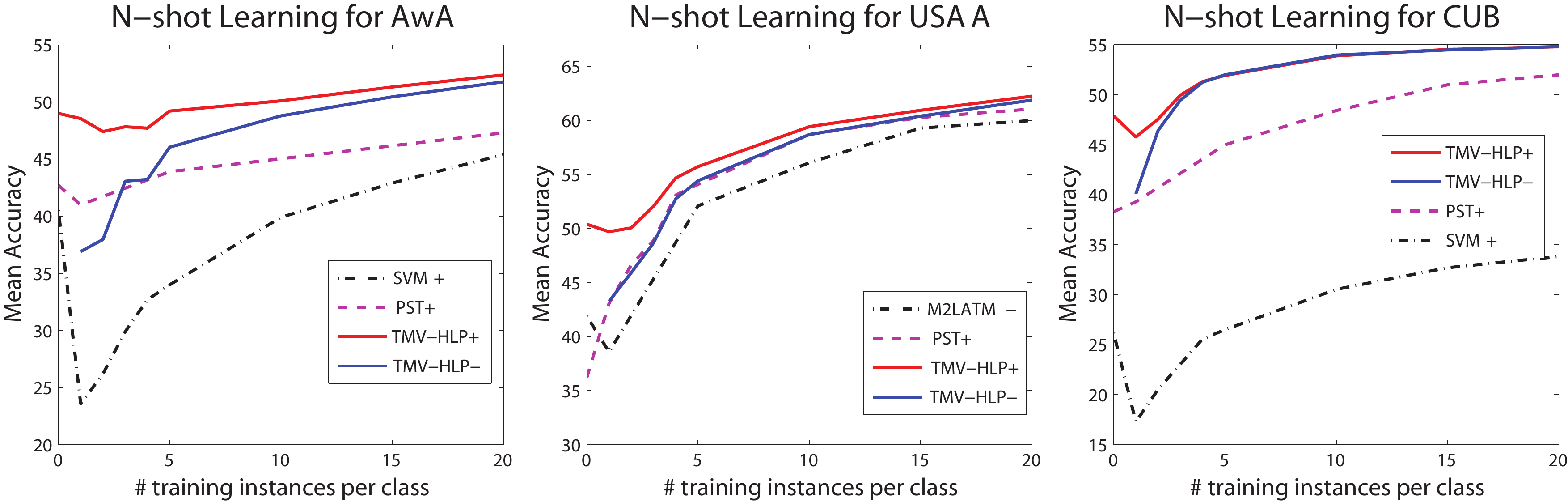} 
\par\end{centering}

\caption{\label{fig:N-shot-learning-for}N-shot learning results with (+) and
without (-) additional prototype information.}
\end{figure*}

\section{N-Shot learning}
\label{sec:n-shot}

N-shot learning experiments are carried out on the three datasets
with the number of target class instances labelled (N) ranging from
0 (zero-shot) to 20. We also consider the situation \cite{transferlearningNIPS}
where both a few training examples \emph{and} a zero-shot prototype
may be available (denoted with suffix $+$), and contrast it to the
conventional N-shot learning setting where solely labelled data and
no prototypes are available (denoted with suffix $-$). For comparison,
PST+ is the method in \cite{transferlearningNIPS} which uses prototypes
for the initial label matrix. SVM+ and M2LATM- are the SVM and M2LATM
methods used in \cite{lampert13AwAPAMI} and \cite{yanweiPAMIlatentattrib}
respectively. For fair comparison, we modify the SVM- used in \cite{lampert13AwAPAMI}
into SVM+ (i.e., add the prototype to the pool of SVM training data).
Note that our TMV-HLP can be used in both conditions but the PST method
\cite{transferlearningNIPS} only applies to the $+$ condition. All
experiments are repeated for $10$ rounds with the average results
reported. Evaluation is done on the remaining unlabelled target data.
From the results shown in Fig.~\ref{fig:N-shot-learning-for}, it
can be seen that: (1) TMV-HLP+ always achieves the best performance,
particularly given few training examples. (2) The methods that explore
transductive learning via label propagation (TMV-HLP+, TMV-HLP-, and
PST+) are clearly superior to those that do not (SMV+ and M2LATM-).
(3) On AwA, PST+ outperforms TMV-HLP- with less than 3 instances per
class. Because PST+ exploits the prototypes, this suggests that a
single good prototype is more informative than a few labelled instances
in N-shot learning. This also explains why sometimes the N-shot learning
results of TMV-HLP+ are worse than its zero-shot learning results
when only few training labels are observed (e.g.~on AwA, the TMV-HLP+
accuracy goes down before going up when more labelled instances are
added). Note that when more labelled instances are available, TMV-HLP-
starts to outperform PST+, because it combines the different views
of the training instances, and the strong effect of the prototypes
is eventually outweighed.

\end{document}